\documentclass{article}

\usepackage{arxiv}

\usepackage[utf8]{inputenc} 
\usepackage[T1]{fontenc}    
\usepackage{hyperref}       
\usepackage{url}            
\usepackage{booktabs}       
\usepackage{nicefrac}       
\usepackage{microtype}      
\usepackage{lipsum}		
\usepackage[sort,compress,numbers]{natbib}
\usepackage{doi}
\usepackage{amsmath,amssymb,amsfonts}
\usepackage{algorithm,algpseudocode}
\usepackage{graphicx}
\usepackage{textcomp}
\usepackage{subcaption}
\usepackage{algpseudocode}
\usepackage[dvipsnames]{xcolor}
\usepackage{multirow}
\usepackage{bibentry}
\usepackage{makecell}
\usepackage{threeparttable}
\usepackage{amssymb}
\usepackage{pifont}
\usepackage{siunitx} 

\title{An Interpretable Neural Control Network with Adaptable Online Learning for Sample Efficient Robot Locomotion Learning}

\author{Arthicha~Srisuchinnawong\\
Vidyasirimedhi Institute of Science and Technology\\Rayong, Thailand, and\\The Mærsk Mc-Kinney Møller Institute\\University of Southern Denmark\\Odense, Denmark.\\
\texttt{arthichas\_pro@vistec.ac.th} \\
\And
Poramate~Manoonpong\\
Vidyasirimedhi Institute of Science and Technology\\Rayong, Thailand, and\\The Mærsk Mc-Kinney Møller Institute\\University of Southern Denmark\\Odense, Denmark.\\
\texttt{poma@mmmi.sdu.dk} \\
}

\renewcommand{\shorttitle}{An Interpretable Neural Control Network with Adaptable Online Learning}

\hypersetup{
pdftitle={A template for the arxiv style},
pdfsubject={q-bio.NC, q-bio.QM},
pdfauthor={David S.~Hippocampus, Elias D.~Striatum},
pdfkeywords={First keyword, Second keyword, More},
}

\begin{document}
\maketitle

\def \ncfull{Sequential Motion Executor}
\def \nc{SME}
\def \ncall{\ncfull{} (\nc)}

\def \lafull{Adaptable Gradient-weighting Online Learning}
\def \la{AGOL}
\def \laall{\lafull{} (\la)}

\def\citenormallearning{\cite{MELA,dogrobot_teacherstudent,kaise_scirobotics,dogrobot_massivelyparallel,LILAC_latent_method,HlifeRL_optionbased,KeepLearning,hexapodlearning_fc_sixmodule,spikingtnnls, hexdog1, dreamwaq, DEAC,viabilityGaitTransition,anymalparkour}}
\def\citealllearning{\cite{MELA,dogrobot_teacherstudent,kaise_scirobotics,dogrobot_massivelyparallel,LILAC_latent_method,HlifeRL_optionbased,KeepLearning,hexapodlearning_fc_sixmodule,fourleg_cpg_multihead,mathias_cpgrbf,hexdog1,spikingtnnls, hexdog1, dreamwaq, DEAC,viabilityGaitTransition}}
\def \citesimplifiedlearning{\cite{mathias_cpgrbf,hexapodlearning_cpg_spikingandforce3leg,fourleg_cpg_multihead}}

\def \figover{Fig.~\ref{fig:neuralcontrol}}
\def \figrewardbatchcpg{Fig.~\ref{fig:cpgbatch}}
\def \figrewardonlinecpg{Fig.~\ref{fig:cpgonline}}
\def \figrewardbatchsme{Fig.~\ref{fig:smebatch}}
\def \figrewardonlinesme{Fig.~\ref{fig:smeonline}}
\def \figresnapa{Fig.~\ref{fig:realsnap1}}
\def \figresnapb{Fig.~\ref{fig:realsnap2}}

\newcommand {\figcurve}[2]{Average episodic rewards, i.e., learning curves and the corresponding min-max range (shade) obtained from five different learning algorithms validated with the #1 neural control architecture #2 under two implementations: (a) batch learning and (b) online learning.}

\def \videolink{\href{https://youtu.be/Svz5H2_MDew}{\url{https://youtu.be/Svz5H2_MDew}}}
\def \videomorfquad{\href{https://youtu.be/Svz5H2_MDew}{\url{https://youtu.be/Svz5H2_MDew}}}
\def \videobone{\href{https://youtu.be/NwAL7jLdTc4}{\url{https://youtu.be/NwAL7jLdTc4}}}
\def \gitlink{ \href{https://github.com/Arthicha/SME-AGOL.git}{\url{github.com/Arthicha/SME-AGOL.git}}}
\def \cpgrbflink{Section~\ref{sec:CPGRBFfigure}}

\newcommand{\na}{\textcolor{gray}{n/a}}
\newcommand{\rd}[1]{\textcolor{red}{#1}}
\newcommand{\mbf}[1]{\mathbf{#1}}
\newcommand{\sbf}[1]{\boldsymbol{#1}}
\newcommand\Tau{\mathrm{T}}
\newcommand\plus{$^\text{+}$}

\newcommand{\boundaryexp}[2]{$C_i$ #1 when #2.}


\newcommand{\review}[1]{#1}
\newcommand{\minorreview}[1]{{\color{blue}#1}}
\newcommand{\mview}[1]{{\color{red}#1}}

\newcommand\lab[1]{\label{#1}}

\begin{abstract}
Robot locomotion learning using reinforcement learning suffers from training sample inefficiency and exhibits the non-understandable/black-box nature. Thus, this work presents a novel \nc-\la{} to address such problems. Firstly, \ncall{} is a three-layer interpretable neural network, where the first produces the sequentially propagating hidden states, the second constructs the corresponding triangular bases with minor non-neighbor interference, and the third maps the bases to the motor commands. Secondly, the \laall{} algorithm prioritizes the update of the parameters with high relevance score, allowing the learning to focus more on the highly relevant ones. Thus, these two components lead to an analyzable framework, where each sequential hidden state/basis represents the learned key poses/robot configuration. Compared to state-of-the-art methods, the \nc-\la{} requires {\SI{40}{\percent}} fewer samples and receives \textbf{\SI{150}{\percent}} higher final reward/locomotion performance on a simulated hexapod robot, while taking merely 10 minutes of learning time from scratch on a physical hexapod robot. Taken together, this work not only proposes the \nc-\la{} for sample efficient and understandable locomotion learning but also emphasizes the potential exploitation of interpretability for improving sample efficiency and learning performance.
\end{abstract}


\section{Introduction}\label{sec:introduction}

Neural networks have demonstrated huge success in a wide range of applications \cite{deeplearning,zumoexo,zumoavis,husky}, including robot locomotion learning \citealllearning. Locomotion Learning typically employs a reinforcement learning technique \cite{RLbook} to optimize the locomotion control parameters based on the reward function and robot-environment interaction. This learning process, therefore, involves a huge quantity of trial-and-error samples, ranging from 150,000 to 400 million timesteps (1 hr--46 days), even on a simple flat terrain \citenormallearning. There are two main causes for this sample inefficiency.

The first cause lies in the neural network control architecture, where thousands of intertwined parameters need to be trained to properly map input observations (i.e., feedback and robot state) to output motor commands for robot gait generation \citenormallearning. Simplifying the architecture can indeed reduce the control parameters and search space. Training a network with two fully connected hidden layers, with nearly 150,000 parameters \cite{MELA}, reportedly takes 5 million timesteps ($\approx$ 5 hrs). Employing a bio-inspired central pattern generator (CPG) as an internal oscillator for basic periodic pattern generation and a radial basis function (RBF) to shape the patterns could reduce the number of parameters to 360 \cite{mathias_cpgrbf}, thus requiring around 160,000 timesteps ($\approx$ 2 hrs) to obtain a stable gait. Sharing the patterns between diagonal legs (i.e., indirect encoding) could further reduce the number of parameters to 120 \cite{mathias_cpgrbf}, potentially accelerating the learning process to 10,000 timesteps ($\approx$ 13 mins). The learning process can be further simplified by predefining intralimb coordination and learning only the control parameters for interlimb coordination. This could involve merely 42 parameters \cite{hexapodlearning_cpg_spikingandforce3leg} and require approximately 66 timesteps to complete ($\approx$ 5 mins). However, robots trained with these simplified strategies could be limited to certain sets of biased actions or gaits (e.g., bilateral symmetric locomotion \cite{mathias_cpgrbf} or identical leg patterns \cite{hexapodlearning_cpg_spikingandforce3leg}), leading to less action/gait diversity and adaptability.

\def \policygradientmaineq{In policy gradient-based learning \cite{RLbook}, a parameter update could be computed from a gradient that maximizes the expected return. \review{This expected return is obtained from an integration, across all possible trajectories, of the product of the probability of each trajectory and its corresponding reward. However, in practical, the expected return must be estimated with a sampling method due to the vast continuous trajectory space and non-differentiable nature of the return.} Consequently, many samples need to be acquired as a batch/set of rollouts before performing the update to stabilize and guide the learning process \citenormallearning.}

The second cause involves the learning algorithm (i.e., the update rule). \policygradientmaineq{} This can cause sample inefficiency and may be impractical in real-world applications. To address this issue, some learning studies applied certain assumptions, such as parameter independence \cite{reinforce,pgpe,ppo} or a unknown input-output mapping function \cite{pibb,hexapodlearning_cpg_spikingandforce3leg}, to simplify the learning rule. Recent locomotion learning approaches have also revisited the use of biased information, such as guided pose \cite{KeepLearning}, to facilitate the optimization process.
 

While the sample efficiency mentioned above remains an unsolved problem, the black-box characteristic of neural networks has been raised as another issue, where a neural network can produce unreliable results \cite{husky}. Therefore, it is essential to equip both the neural network architecture and learning algorithm with interpretability, allowing developers and users to understand and analyze them effectively \cite{reviewInterpretableRL,husky,neurovis}. As a result, designing decomposable modular neural architectures has been suggested as a solution to obtain an interpretable neural architecture \cite{reviewInterpretableRL}, where the network could be divided into modules \cite{zumoexo,zumoavis,mathias_nature,pbird_morfendocrine} or two-neuron design operators \cite{interpRL_dooperation} with specific functionalities. Furthermore, post-hoc explanation techniques, such as local interpretable model-agnostic explanation (LIME) \cite{husky}, layer-wise relevance propagation (LRP) \cite{lrp}, and guided backpropagation (GBP) \cite{gbp}, have been proposed for unveiling learning algorithms and their results through network attention and key feature interpretation.

Although the locomotion learning performance and interpretability seem to be two different unrelated aspects, we hypothesize that interpretability could also be utilized to facilitate and speed up locomotion learning. Thus, this study proposes efficient neural control and learning mechanisms combining these two aspects. Starting from an interpretable \ncall{} architecture (i.e., an interpretable neural control network), each neuron is designed to serve specific functions (i.e., internal state and key pose\footnote{Here, a key pose refers to a specific robot configuration or a set of joint positions.}), providing less interference between non-related activities to ease the collective learning \cite{RLbook}, as discussed in Section~\ref{sec:neuralcontrol}. An \lafull{}
algorithm (\la) (i.e., an adaptable online learning rule) is also developed. It can be adapted and scaled according to relevance/importance scores \cite{lrp,gbp}, allowing the learning to focus more on highly relevant control parameters, as discussed in Section~\ref{sec:learning}. 

In summary, the contributions of this work include:

\begin{enumerate}
	\item An interpretable \ncall{} network, along with its design principle, serving as a foundational structure to achieve sample efficiency and enhance locomotion learning.
	\item An \laall{} algorithm, acting as an online control parameter optimizer to achieve sample efficiency and enhance locomotion learning.
\end{enumerate}

In addition to the key contributions, this study also demonstrates the performance of the proposed \nc-\la{} mechanisms in hexapod robot locomotion learning in the real world. This demonstration highlights the potential of leveraging interpretability to improve both the sample efficiency and performance of robot locomotion learning.

\label{sec:morf}

\def \morfdiagramcap{\review{The physical and simulated versions of MORF including the motor positions of a leg, their rotational axes, the RealSense tracking camera, the world frame, the robot frame, and the transformation between the world and robot frames.}}


\def\morfdescribe{\review{To verify the proposed \nc-\la{} mechanisms, this study employs a hexapod robot (MORF) \cite{morf} with six legs (Fig.~\ref{fig:morf}), each of which has three active joints: right front leg ($RF_{1-3}$), right middle leg ($RM_{1-3}$), right front leg ($RH_{1-3}$), left front leg ($LF_{1-3}$), left middle leg ($LM_{1-3}$), and left front leg ($LH_{1-3}$). Each joint is controlled by a Dynamixel servo motor, receiving target position commands as input. For sensory feedback, here the robot utilizes two modalities: motor position feedback for low-level PID control and visual odometry feedback for estimating its traveled distance. The traveled distance estimation is used for reward computation of the AGOL. The odometry feedback is obtained from an Intel Realsense tracking camera T265 in the case of a physical robot and computed from the global robot position and orientation in the case of simulation (Coppeliasim \cite{vrep} with Mujoco physics engine \cite{mujoco}).}}

\def \smeagoloverview{\review{In this section, we present the \nc-\la{} architecture (Fig.~\ref{fig:neuralcontrol}), which consists of the \ncfull{} mechanism (\nc, described in Section~\ref{sec:neuralcontrol}) and the \lafull{} mechanism (\la, described in Section II-B). The \nc{} acts as neural control for locomotion generation, while the \la{} functions as an online learning algorithm for adapting the control parameters.}}

\section{Interpretable \ncfull-\lafull{} (\nc-\la)}
\label{sec:method}

\def \ncdiagramcap{An overview of the \ncfull-\lafull{} (\nc-\la{}) architecture, presented along with the corresponding signals \review{obtained from three layers, and the experimental hexapod robot platform MORF \cite{morf} for the study. The three layers includes a central pattern generator layer ($c_i$, red) providing the discrete (non-smooth) robot states, a basis layer ($b_i$, green) providing the smooth version of the robot states or the movement bases, and an output layer (\{$M_i$\} = \{$RF_{1-3}$, $RM_{1-3}$, $RH_{1-3}$, $LF_{1-3}$, $LM_{1-3}$, $LH_{1-3}$\}) providing the motor commands (e.g., $RF_{1-3}$, blue), which can be interpreted as the interpolated key poses.}}

\begin{figure}[!h]
	\centering
	\begin{subfigure}{0.7\linewidth}
		\includegraphics[width=\linewidth]{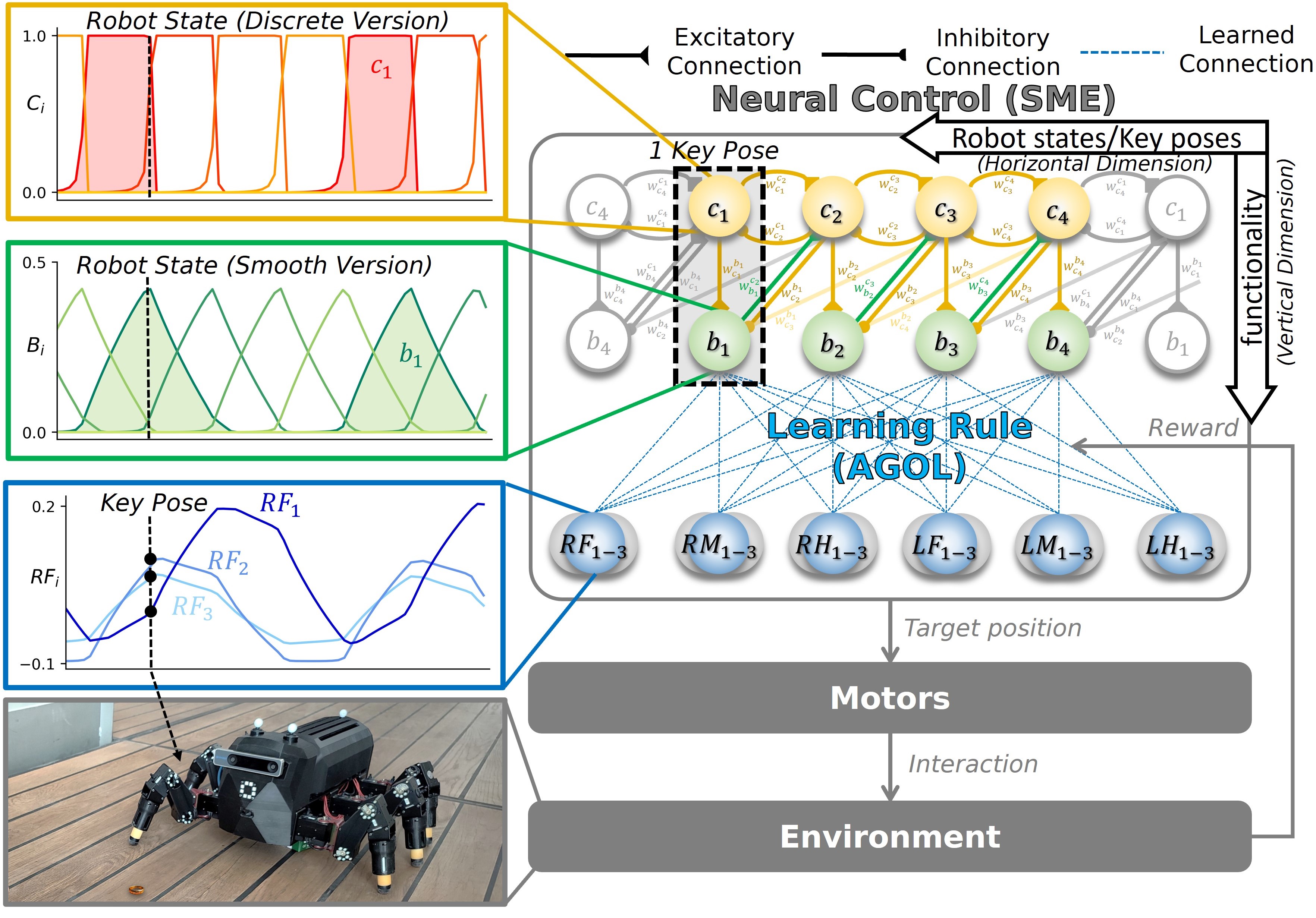}
		\caption{}
		\label{fig:neuralcontrol}
	\end{subfigure}
	\\ 
	\centering
	\begin{subfigure}{0.7\linewidth}
		\includegraphics[width=0.47\linewidth]{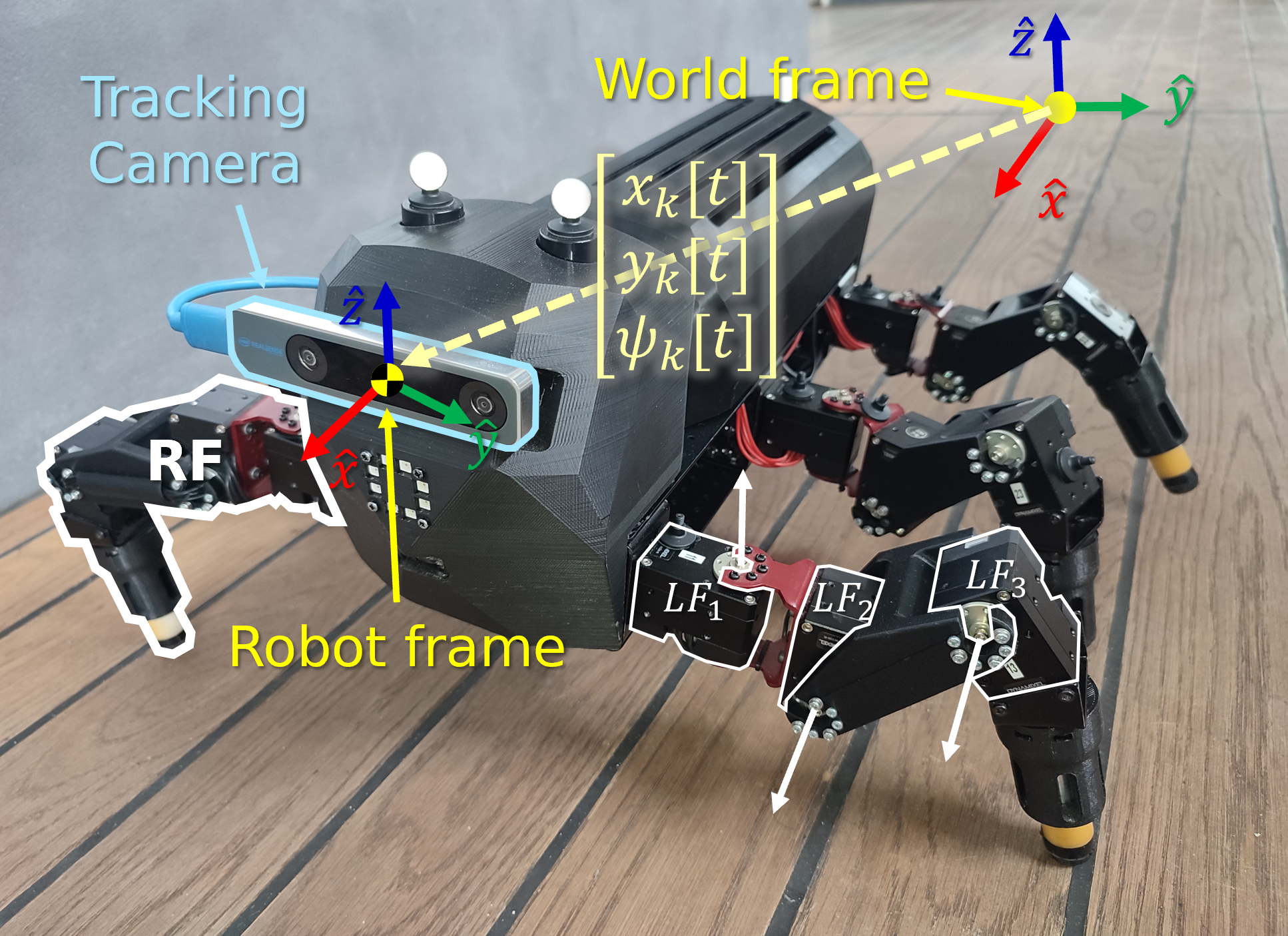}
		\hfill
		\includegraphics[width=0.47\linewidth]{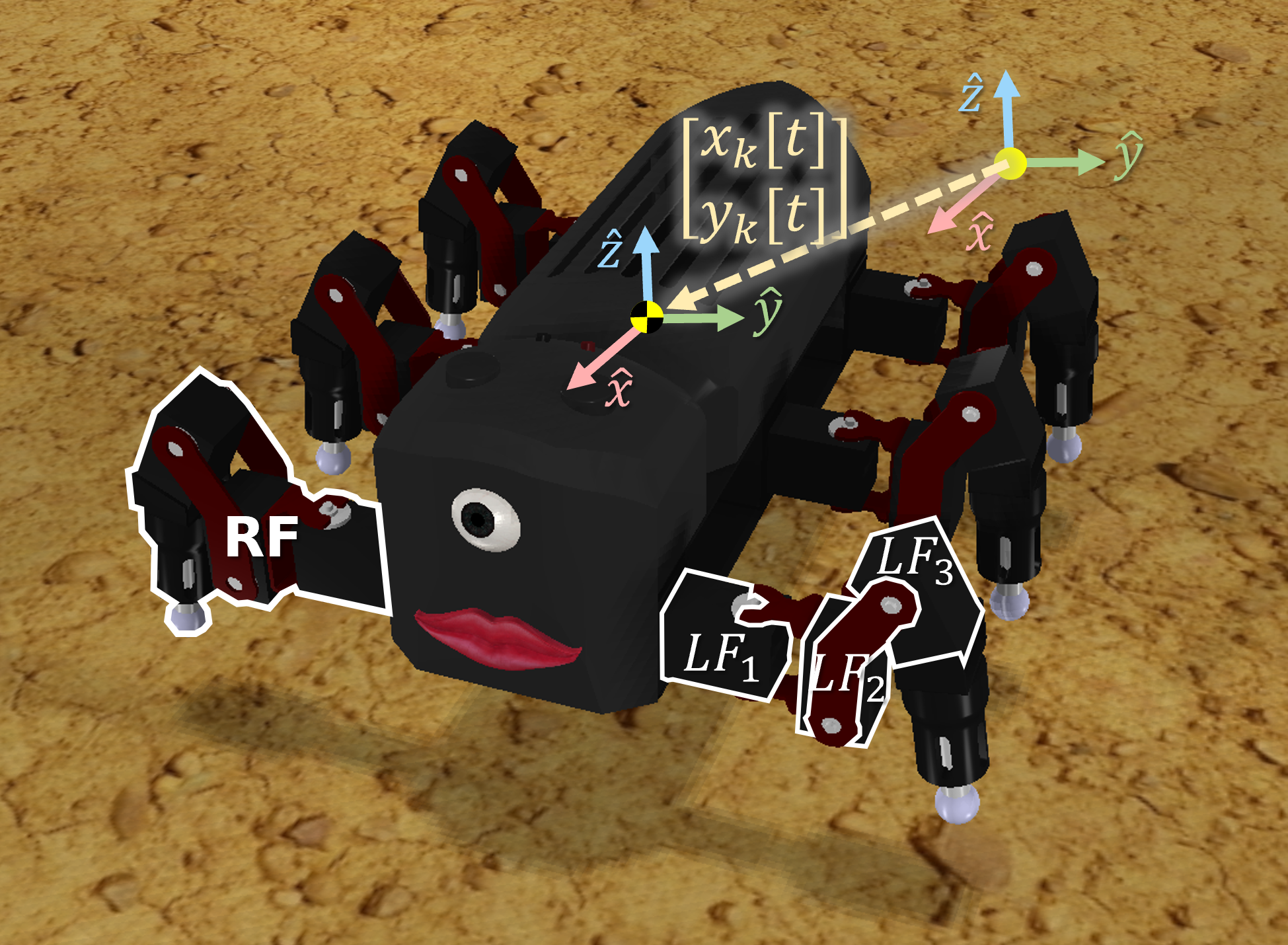}
		\caption{}
		\label{fig:morf}
	\end{subfigure}
	\caption{(a) \ncdiagramcap{} (b) \morfdiagramcap}%
	\label{fig:neuralcontrolandmorf}
\end{figure}

\smeagoloverview

\morfdescribe

\subsection{\ncall{} Neural Control}\label{sec:neuralcontrol}

\def \interpneuralcontrol{\review{The SME (Fig.~\ref{fig:neuralcontrol}) is an interpretable, discrete-time neural control network. It generates target motor position commands based solely on its internal states. The motor commands drive the robot's locomotion. The control network is designed with two dimensions of interpretability: vertical and horizontal dimensions. The vertical dimension reflects the overall control functionality, while the horizontal dimension encodes the robot's states or key poses\footnote{Here, a key pose refers to a specific robot configuration or a set of joint positions.} (robot motion encoding).} 
	
	\review{Along the vertical dimension,} the network consists of three neuron layers: central pattern generator neurons (Cs), basis neurons (Bs), and output neurons (RFs--LHs). Cs \review{provide the central patterns, which} are pre-configured to activate sequentially, as depicted in red/yellow \review{(the top left plot in Fig.~\ref{fig:neuralcontrol})}. These neural activities are subsequently fed to Bs in the basis layer to produce triangle-shaped basis signals, as depicted in green \review{(the middle left plot in Fig.~\ref{fig:neuralcontrol})}. Accordingly, these Bs are also activated in the same order as Cs; however, they only overlap with their neighbors, providing less interference between non-neighbor bases while maintaining smooth transitions. Finally, the basis activities are mapped to the output motor commands, as depicted in blue \review{(the bottom left plot in Fig.~\ref{fig:neuralcontrol})}, through the learned connections.

	\review{Along the horizontal dimension, each column-wise alignment represents a specific robot state corresponding to a key pose (the black dashed square in Fig.~\ref{fig:neuralcontrol}). In this network structure, each column has a central pattern generator neuron that generates a basic, discrete state. This state is then further shaped by the basis neuron into a smoother state. The learned connections (the blue dashed lines in Fig.~\ref{fig:neuralcontrol}) encode the joint/motor position commands corresponding to the robot states, where the one from $b_i$ to $M_i$ represents the key pose of joint $M_i$ when $b_i$ fully activates, and $M_i$ denotes the output neurons (i.e., $RF_{1-3}$, $RM_{1-3}$, $RH_{1-3}$, $LF_{1-3}$, $LM_{1-3}$, and $LH_{1-3}$).} It should be noted that, in this work, the entire network is updated at 20 Hz, with four sets of neurons ($\{c_i, b_i\}$, $i \in \{1, 2, 3, 4\}$), \review{representing minimal four key poses:} two for the swing phase and two for the stance phase, as presented in Fig.~\ref{fig:neuralcontrol}.}

\interpneuralcontrol

\subsubsection{Central Pattern Generator Neurons (Cs)}
Functioning as the internal states of the network, Cs use the corresponding bases to trigger the propagation to the next. The activity of each neuron is governed by:
\begin{equation}
\mbf{c}[\mathbin{{t}{+}{1}}] = \text{sigmoid} \left(  \mbf{W}^c_c \mbf{c}[t]  + \mbf{W}^c_b \mbf{b}[t] + b^c  \right),
\label{eq:zpg}
\end{equation}
where $\mbf{c}[t]$ and $\mbf{b}_i[t]$ denote the activation vector of the central pattern generator neurons at timestep $t$ ($c_i[t]$) and that of the bases ($b_i[t]$), $\mbf{W}^c_c$ and $\mbf{W}^c_b$ denote the connection weight matrices from Cs to Cs and Bs to Cs, respectively (it should be note that their elements at row $i^{\text{th}}$ and column $j^{\text{th}}$ are denoted by $w^{c_i}_{c_j}$ and $w^{c_i}_{b_j}$, respectively), and $b^c$ denotes a scalar bias. 

The parameters $w^{c_i}_{c_p}$, $w^{c_i}_{c_i}$, $w^{c_i}_{c_n}$, $w^{c_i}_{b_p}$ and $b^c$, where $p$, $i$, and $n$ denote the index of the previous, current, and next neurons in the sequence, are selected analytically such that $c_i[t]$ should:
\begin{itemize}
	\item be $\gamma$ when the previous state and previous basis neurons are both active ($c_p[t] = b_p[t] = \iota$) while the others remain inhibited,
	\item remain inhibited around $-\omega$ when only the previous state is active ($c_p[t] = \iota$),
	\item remain inhibited around $-\omega$ when only the previous basis is active ($b_p[t] = \iota$),
	\item be maintained around $\omega$ when it is previously active ($c_i[t] = \iota$),
	\item be inhibited to around $-\omega$ when the next internal state is active ($c_n[t] = \iota$) even if others remain active.
\end{itemize}
 These yield a system of boundary equations as:
\begin{equation}
\begin{bmatrix}
\gamma \\ -\omega \\ -\omega \\ \omega \\ -\omega 
\end{bmatrix} 
=
\begin{bmatrix}
\iota & \epsilon & \epsilon & \iota & 1 \\
\iota & \epsilon & \epsilon & \epsilon & 1 \\
\epsilon & \epsilon & \epsilon & \iota & 1 \\
\epsilon & \iota & \epsilon & \epsilon & 1 \\
\iota & \iota & \iota & \iota & 1 \\
\end{bmatrix} 
\begin{bmatrix}
w^{c_i}_{c_p} \\ w^{c_i}_{c_i} \\ w^{c_i}_{c_n} \\ w^{c_i}_{b_p} \\ b^c
\end{bmatrix},
\label{eq:cparam}
\end{equation}
where $\gamma$ refers to the minimum activity of an active neuron, $\omega$ refers to a value corresponding to $\text{sigmoid}(\omega) \rightarrow 1$, $\iota$ refers to fully excited activity, and $\epsilon$ refers to fully inhibited activity. 

In this work, the free parameters $\gamma$, $\omega$, $\iota$, and $\epsilon$ are chosen as 0.5, 8, 0.95, and 0.01, respectively. This yields $w^{c_i}_{c_p}$, $w^{c_i}_{c_i}$, $w^{c_i}_{c_n}$, $w^{c_i}_{b_p}$ and $b^c$ of 8, 25, -32, 8, and -15. With these parameters, the resulting four central pattern generator outputs ($c_{1-4}[t]$) are illustrated in \figover{} (radish/yellowish). These output signals are then fed to the basis neurons to produce bases for output mapping.

\subsubsection{Basis Neurons (Bs)}
Receiving the internal states/central pattern generator outputs, the basis neurons are formulated as Eq.~\ref{eq:basis} to produce triangular periodic signals, interpreted as key poses (sets of all joint positions). 
\begin{equation}
\mbf{b}[\mathbin{{t}{+}{1}}] =\text{ReLU} \left(  \mbf{W}^b_c \mbf{c}[t]  + \mbf{W}^b_b \mbf{b}[t]  \right),
\label{eq:basis}
\end{equation}
where $\mbf{c}[t]$ and $\mbf{b}_i[t]$ denote the activation vector of the central pattern generator outputs at timestep $t$ ($c_i[t]$) and that of the bases ($b_i[t]$), $\text{ReLU}()$ denotes the \review{rectified} linear unit function, and $\mbf{W}^b_c$ and $\mbf{W}^b_b$ denote the connection weight matrices from Cs to Bs and Bs to Bs, respectively (their elements are denoted by $w^{b_i}_{c_j}$ and $w^{b_i}_{b_j}$).

\begin{figure}[!h]
	\centering
	\begin{minipage}{0.6\linewidth}
		\begin{subfigure}{\linewidth}
			\includegraphics[width=\linewidth]{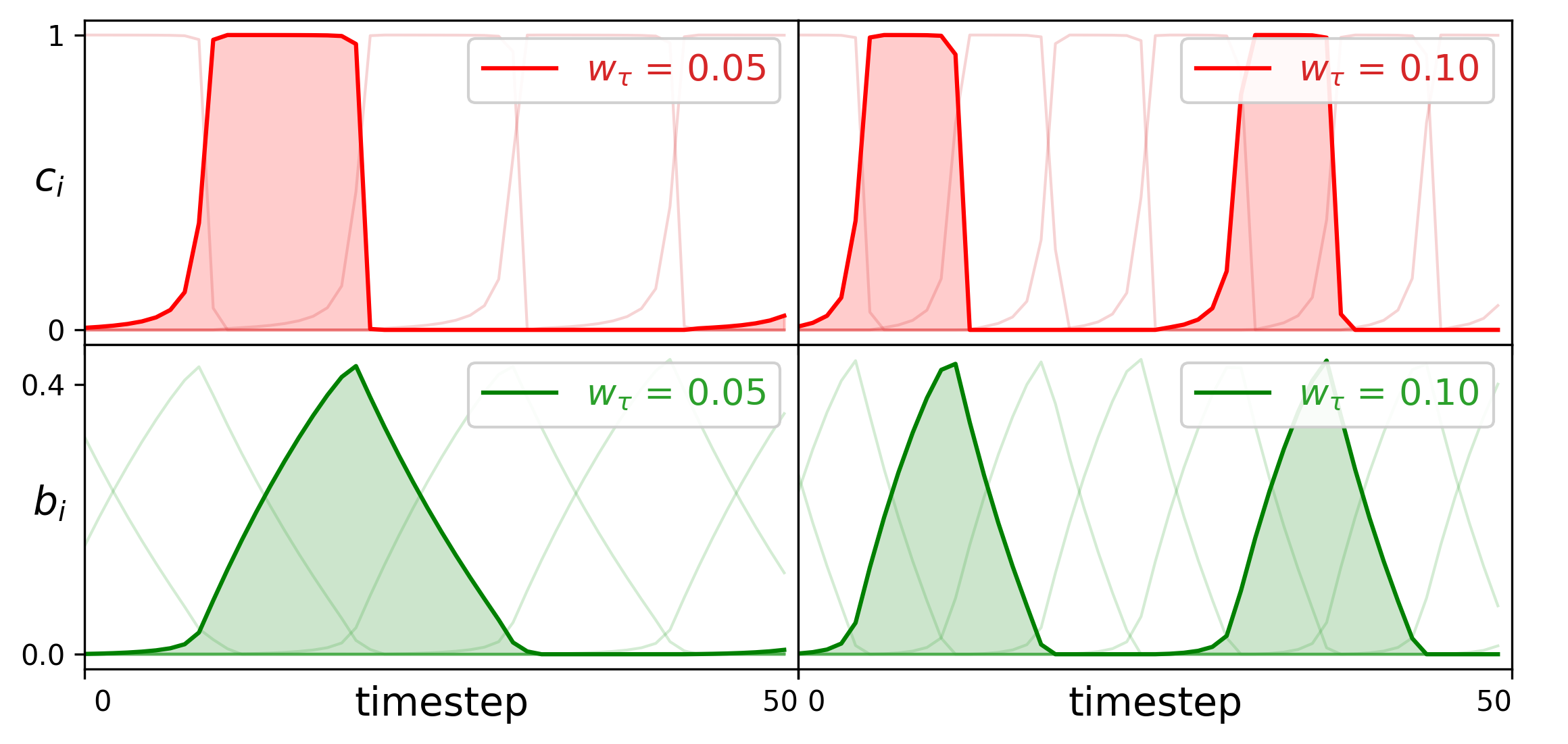}
			\caption{}
			\label{fig:basis_freq}
		\end{subfigure} \\
		\medskip
		\begin{subfigure}{\linewidth}
			\includegraphics[width=\linewidth]{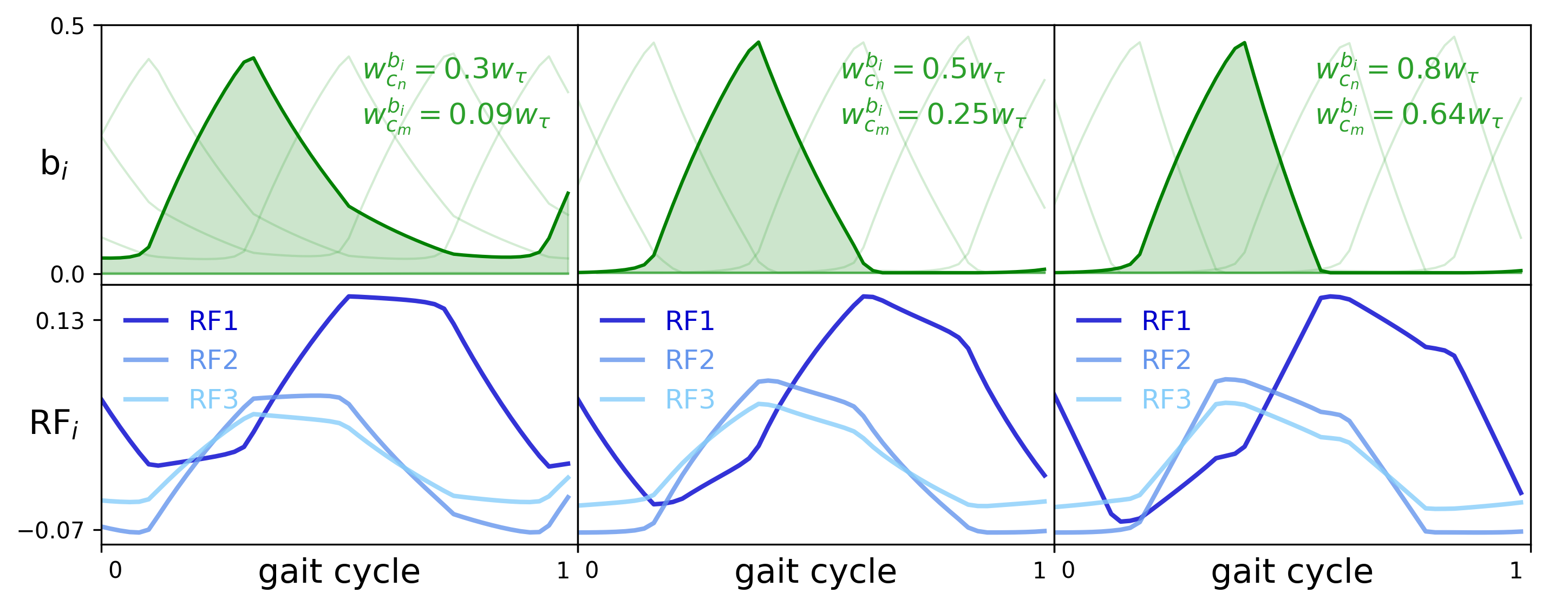}
			\caption{}
			\label{fig:basis_shape}
		\end{subfigure}  
	\end{minipage} 
	\hfill
	\begin{minipage}{0.35\linewidth}
		\begin{subfigure}{\linewidth}
			\includegraphics[width=\linewidth]{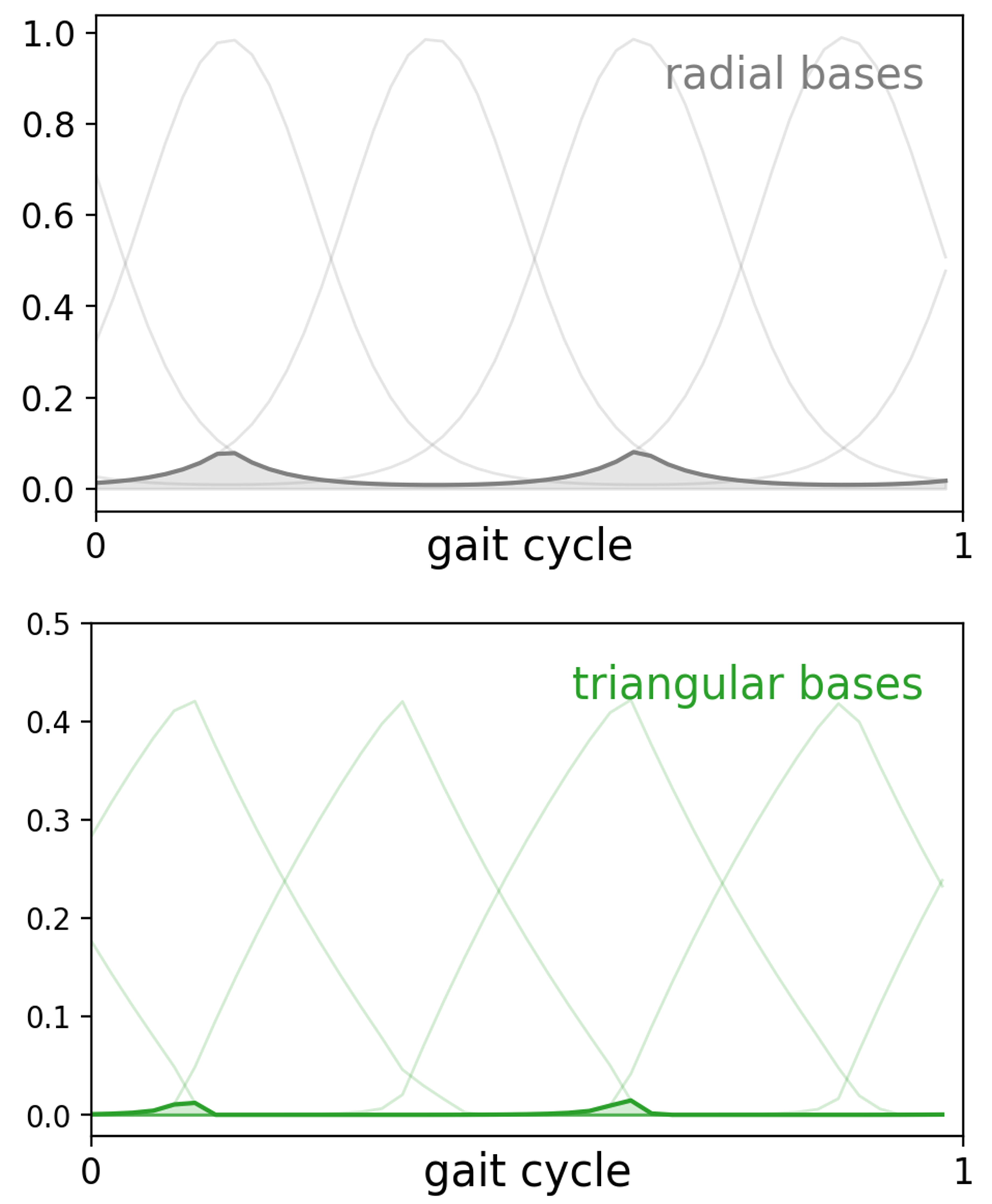}
			\caption{}
			\label{fig:basis_vs_rbf} 
		\end{subfigure} 
	\end{minipage}

	\caption{(a) Central pattern generator neurons/internal states ($c_i$) and bases ($b_i$) obtained from \nc{} where $w_\tau$ = (left) 0.05 and (right) 0.10. (b) Intersection between two non-neighbor (left) radial bases and (right) equivalent triangular bases. (c) Bases ($b_i$) and outputs ($RF_i$) obtained from \nc{} where $w^{b_i}_{c_n}$ and $w^{b_i}_{c_m}$ are set as (left) 0.3$w_\tau$ and 0.09$w_\tau$, (middle) 0.5$w_\tau$ and 0.25$w_\tau$, and (right) 0.8$w_\tau$ and 0.64$w_\tau$.}%
	\label{fig:basis}
\end{figure}

The parameters $w^{b_i}_{c_i}$ and $w^{b_i}_{b_i}$, are selected as $w_\tau$ and $1-w_\tau$ such that each basis neuron performs lowpass filtering of the corresponding central pattern generator neuron/internal state. Thus, $w_\tau$ is another free parameter, controlling the transition speed/walking frequency, as illustrated in Fig.~\ref{fig:basis_freq}. In this work, $w_\tau$ is selected as 0.05 to obtain a gait frequency of 0.3 Hz.

\def \pthreethree{With this setup, the resulting triangular bases have \review{\SI{33}{\percent}} less intersection between non-neighbor bases (\review{\SI{7}{\percent}} of the time, \review{p-value $<$ 0.05, Mann-Whitney U test}) than those of equivalent radial bases employed in \cite{mathias_cpgrbf} (\review{\SI{40}{\percent}} of the time), as shown in Fig.~\ref{fig:basis_vs_rbf}.}

The parameters $w^{b_i}_{c_n}$ and $w^{b_i}_{c_m}$, where $n$ and $m$ denote the index of the next and second neurons in the sequence, are then selected empirically, as illustrated in Fig.~\ref{fig:basis_shape}, to shape the bases. $w^{b_i}_{c_n}$ and $w^{b_i}_{c_m}$ are set as 0.5$w_\tau$ and 0.25$w_\tau$, respectively, to reduce interference/intersection between non-neighbors (Fig.~\ref{fig:basis_shape}-left and Fig.~\ref{fig:basis_vs_rbf}) and discontinuous outputs (Fig.~\ref{fig:basis_shape}-right). \pthreethree{} At each timestep, these bases are then mapped to the outputs through learned connections, where less interference can reduce the number of active bases, thereby decreasing the number of relevant parameters/connections weights that need to be updated. Consequently, this mechanism can identify the important control parameters (here, output mapping weights), facilitating the learning process. In other words, it encourages the learning process to focus more on a few highly relevant control parameters at each time/iteration, as shown in Fig.~\ref{fig:basis_vs_rbf} when compared with CPGRBF \cite{mathias_cpgrbf}.

\subsubsection{Output Neurons (RFs--LHs)}

In this work, a learned weight matrix $\mbf{W}^o_b$ (18 rows/outputs $\times$ 4 columns/key poses) translates the bases to 18 outputs, employed as motor position commands (i.e., direct encoding \cite{mathias_cpgrbf}). The outputs are computed according to:
\begin{equation}
\mbf{o}[\mathbin{{t}{+}{1}}] = \text{clip} \left(  \mbf{W}^o_b \mbf{b}[t]  , -\alpha_{\text{max}},  \alpha_{\text{max}} \right),
\label{eq:output}
\end{equation}
where $\mbf{o}[t]$ denotes the output vector $t$ ($RF_1[t]$--$LH_3[t]$), $\mbf{b}_i[t]$ denotes the activation vector of the bases at timestep $t$ ($b_i[t]$), $\text{clip}()$ denotes the saturation function, ensuring that the outputs are between the joint limit ($-\alpha_{\text{max}}$,  $\alpha_{\text{max}}$). $\alpha_{\text{max}}$ is 0.3 in this work, and $\mbf{W}^o_b$ denotes the connection weight matrices from Bs to the outputs.

\def \legcoorlearning{\review{As each $b_i$ neuron is activated one after another, they can be interpreted as the activation of certain key poses. Four basis activities ($b_{1-4}$) are then mapped to 12 motor positional commands via the learned weights/connections between the basis layer ($b_i$s) and the output layer (RF--LH). As these 4 $\times$ 12 weight values are learned independently (i.e., direct encoding \cite{mathias_cpgrbf}), the robot can learn/adjust each leg patterns independently. Therefore, inter-leg coordination is achieved by adjusting/learning these 4 $\times$ 12 mapping weight values, as shown in Sections~\ref{sec:legcoordination} and \ref{sec:differentshape} in the supplementary material. Leg coordination pattern can be obtained through swapped $\mbf{W}^o_b$ weight values, as shown in Figs.~\ref{fig:legco0}--\ref{fig:legco180} in Section~\ref{sec:legcoordination} in the supplementary material, while different shape signals can be obtained through different $\mbf{W}^o_b$ weight values, as shown in Fig.~\ref{fig:legshape} in Section~\ref{sec:differentshape} in the supplementary material. In this work, the weight matrix $\mbf{W}^o_b$ was trained with the \la{} to achieve leg coordination and different leg patterns automatically based in the rewards obtained from trial-and-error, as described in the following section.}}

\legcoorlearning

\subsection{\laall}\label{sec:learning}
During the training, a sampling method is applied along with the assumption of a deterministic environment ($\nabla_\theta \ln p(s_0) = 0$ and $\nabla_\theta \ln p(s_{t+1}|s_t,\tilde{a}_t) = 0$), yielding:
\begin{align}
\Delta \theta
&\approx \eta_\theta \sum_\tau \nabla_{\theta} \ln{ \left(  p(s_0) \prod_i p(s_{t+1}|s_t,\tilde{a}_t) p(\tilde{a}_t|\theta ) \right) }   R_t , \notag \\
&\approx \eta_\theta \sum_\tau \sum_{t} \nabla_{\theta} \ln{ p(\tilde{a}_t|\theta )  }  \; R_t, \label{eq:gradgeneral} 
\end{align}
where $\Delta \theta$ denotes the weight update, $\eta_\theta$ denotes the learning rate, $\tau$ denotes a trajectory, $p(s_0)$ denotes the probability distribution of being at \review{the} state $s_0$, $p(s_{t+1}|s_t,\tilde{a}_t)$ denotes the probability distribution of being at $s_{t+1}$ given the previous state $s_t$ and explored action $\tilde{a}_t$, $p(\tilde{a}_t|\theta)$ denotes the probability distribution of performing explored action $\tilde{a}_t$ given network parameter $\theta$, and $R_t$ denotes the corresponding return computed from the summation of all the rewards obtained ($R = \sum_{t'} r_{t'}$).

In action space exploration learning algorithms, like policy gradient (PG) \cite{reinforce}, $p(\tilde{a}_t|\theta)$ is estimated with a normal distribution of explored action $\tilde{a}_t$ given actual differentiable action $a_t$ (i.e., $\mathcal{N}(\tilde{a}_t;a_t,\sigma_a)$ or $\pi_\theta(\tilde{a}_t|a_t)$), while proximal policy optimization (PPO) \cite{ppo} further incorporates importance sampling between old and new policies. However, \cite{exploration_parameter_vs_action,pgpe,parameterspace_exploration} reported that exploring policy in the parameter space outperforms the action space counterparts due to consistency. As a result, a policy gradient with parameter-based exploration (PGPE) \cite{pgpe} revises the learning rule in Eq.~\ref{eq:gradgeneral} by performing exploration in the parameter space while assuming $\nabla_\theta \ln p(\tilde{a}_t|\tilde{\theta}_t) = 0$ due to the deterministic relationship \review{between $\tilde{a}_t$} and $\tilde{\theta}_t$. This yielded:
\begin{align}
\Delta \theta
&\approx \eta_\theta \sum_\tau \sum_t  \nabla_{\theta} \ln \left( p(\tilde{a}_t|\tilde{\theta}_t ) p(\tilde{\theta}_t|\theta) \right)  R_t , \notag \\
&\approx \eta_\theta \sum_\tau \sum_t \nabla_{\theta} \ln \mathcal{N}(\tilde{\theta}_t;\theta,\sigma_\theta)     \; R_t , \notag \\
&\approx \eta_\theta \sum_\tau \sum_t  \frac{(\tilde{\theta}_t-\theta)}{\sigma_\theta^2}  \; R_t  , 
\label{eq:gradparam}
\end{align} 
where $\Delta \theta$ denotes the weight update, $\eta_\theta$ denotes the learning rate, $\tau$ denotes a trajectory, $p(\tilde{a}_t|\tilde{\theta}_t )$ denotes the probability distribution of explored action $\tilde{a}_t$ given explored parameter $\tilde{\theta}_t$, $p(\tilde{\theta}_t|\theta) $ denotes the probability distribution of explored parameter $\tilde{\theta}_t$ given parameter $\theta$ and standard deviation $\sigma_\theta$ ($\mathcal{N}(\tilde{\theta}_t;\theta,\sigma_\theta)$), and $R_t$ denotes the return.

From Eq.~\ref{eq:gradparam}, in the policy gradient with parameter-based exploration (PGPE) \cite{pgpe}, $R_t$ is replaced with the difference between the obtained reward and estimated baseline (advantage, $A_t$), whereas,  \review{in the policy improvement} with black-box optimization (PIBB) \cite{pibb}, it is substituted with the ranking score. While these approaches assume that $\nabla_\theta \ln p(\tilde{a}_t|\tilde{\theta}_t) = 0$, the parameter $\theta$ \review{is} updated even if it does not contribute to the explored action $\tilde{a}_t$. 

\def\learningdiagramcap{\review{Online locomotion learning  process. Firstly, the robot interacts with the environment through the use of explored action $\tilde{a}_t$ generated from the \nc{} neural control. Secondly, the \la{} learning rule updates the \nc{} control parameters using the samples from the trajectory $\tau$, consisting of the reward obtained from the interaction with environment $r_t$, the network state (including the parameters $\theta$, explored parameters $\tilde{\theta}_t$, and exploration standard deviation $\sigma_\theta$), and the relevance explanation $|\text{Rel}_{\tilde{\theta}}|$ computed from the network state.}}
\begin{figure}[!h]
	\centering
	\includegraphics[width=0.6\linewidth]{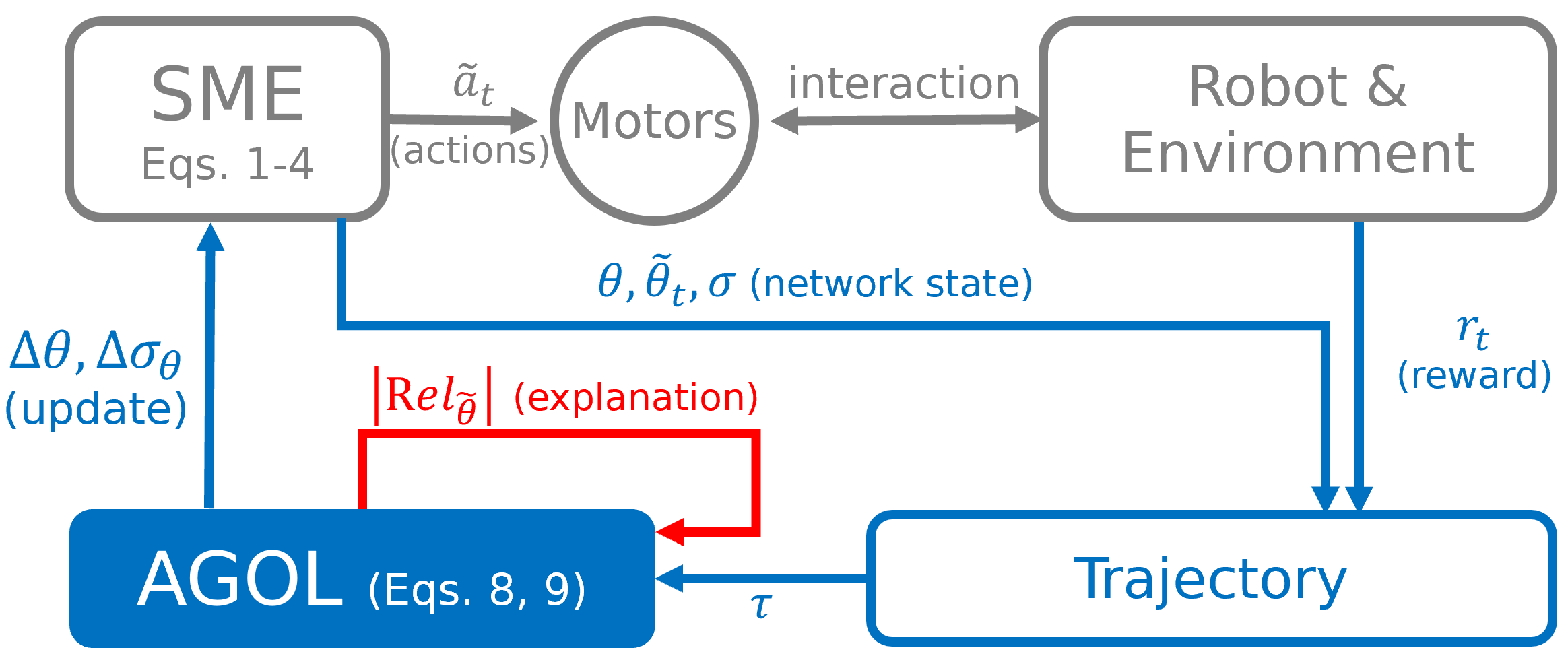}
	\caption{\learningdiagramcap}
	\label{fig:learningconcept}
\end{figure}

Therefore, as an improvement, the learning rule in Eq.~\ref{eq:gradparam} can be modified by introducing the (absolute) relevance of the parameter ($|\text{Rel}_{\tilde{\theta}}|$), according to: 
\begin{align}
\Delta \theta 
&\approx \eta_\theta \sum_\tau \sum_t  |\text{Rel}_{\tilde{\theta}}| \frac{(\tilde{\theta}_t-\theta)}{\sigma_\theta^2}  \; A_t, \label{eq:agol_rel} 
\end{align} 
where $\Delta \theta$ denotes the weight update, $\eta_\theta$ denotes the learning rate, $\tau$ denotes a trajectory, $\tilde{a}_t$ denotes the explored action according to the explored parameter $\tilde{\theta}_t$, $\sigma_\theta$ denotes the exploration standard deviation, $\theta$ denotes the actual network parameter, and $A_t$ denotes the corresponding advantage estimated from the reward/return $R_t$ and the baseline (e.g., average reward). 

To exploit PyTorch autograd \cite{pytorch} for efficient computation, this work substitutes $\text{Rel}_{\tilde{\theta}}$ with the gradient of the action with respect to the parameters ($|\nabla_{\tilde{\theta}} \tilde{a}_t|$), according to:
\begin{align}
\Delta \theta 
&\approx \eta_\theta \sum_\tau \sum_t  |\nabla_{\tilde{\theta}} \tilde{a}_t| \frac{(\tilde{\theta}_t-\theta)}{\sigma_\theta^2}  \; A_t, \label{eq:agol}
\end{align} 
where $\Delta \theta$ denotes the weight update, $\eta_\theta$ denotes the learning rate, $\tau$ denotes a trajectory, $\tilde{a}_t$ denotes the explored action according to the explored parameter $\tilde{\theta}_t$, $\sigma_\theta$ denotes the exploration standard deviation, $\theta$ denotes the actual network parameter, and $A_t$ denotes the corresponding advantage estimated from the reward/return $R_t$ and the baseline (e.g., average reward).

Using this learning rule, at each $t$, the output gradient cannot be backpropagated to the parameters that do not contribute to the outputs/actions, resulting in $|\nabla_{\tilde{\theta}} \tilde{a}_t| \rightarrow 0$. Moreover, the added gradient upweights the updates of more relevant parameters and downweights those of less relevant parameters. Intuitively, this learning rule equips robots with the ability to explain the relevance of each parameter to itself and scale the update accordingly, as illustrated in Fig.~\ref{fig:learningconcept}, allowing the robot to focus on a few important parameters each time.

Following a similar procedure, the exploration weights can also be adapted online ($\Delta \sigma_\theta$) using the gradient with respect to the exploration standard deviation ($\nabla_{\sigma_\theta} \ln{p(\tau|\theta)}$) and an exploration adaptation/learning rate ($\eta_\sigma$), yielding:
\begin{align}
\Delta \sigma_\theta 
&\approx \eta_\sigma \sum_\tau \sum_t \left( |\nabla_{\tilde{\theta}} \tilde{a}_t| \frac{ (\tilde{\theta}_t-\theta)^2-\sigma_\theta^2 }{\sigma_\theta^3}  A_t \right).
\label{eq:agol_sigma}
\end{align} 

In this work, to learn hexapod locomotion, this learning rule is applied to update the connection weights ($\mbf{W}^o_b$), mapping the bases to outputs, all initialized as zero. Two variants of \la{} are also studied. The first (\la) performs online learning by updating the parameter after each episode ends, while the second (Added Gradient Learning or AGL) performs batch learning by updating the parameter after certain episodes (here, eight episodes are sufficient for batch learning as shown in \cite{mathias_cpgrbf}).

\section{Experiments and Results} \label{sec:exp}
To verify the proposed \nc-\la{}, two experiments were conducted. The first experiment (Section~\ref{sec:expsim}) compared five different state-of-the-art learning algorithms, cross-implemented with two implementations (batch and online learning) and two neural control architectures, on a simulated hexapod robot, where reward and sample efficiency (learning speed) were selected as the comparison metrics. The second experiment (Section~\ref{sec:expreal}) then extended this to a physical robot to validate the \nc-\la{} performance in the real world.

\subsection{Simulation Experiment}
\label{sec:expsim}

In this experiment, two neural control architectures (\nc{} and the state-of-the-art control (CPGRBF) \cite{mathias_cpgrbf} \review{in \cpgrbflink{} of the supplementary document}) were compared due to their comparable mapping process (i.e., radial bases in CPGRBF versus triangular bases in \nc) and having the identical parameter numbers (i.e., 72 parameters). For the learning, two action space exploration learning algorithms (policy gradient (PG) \cite{reinforce} and proximal policy optimization (PPO) \cite{ppo}) along with three parameter space exploration learning algorithms (policy improvement with black-box optimization (PIBB) \cite{pibb}, policy gradient with parameter-based exploration (PGPE) \cite{pgpe}, and the proposed \la{}), were cross-implemented with the control architectures. For each learning algorithm, batch learning (Algorithm \ref{alg:batch}) and online learning (Algorithm \ref{alg:online}) were also included. For clarity, $^+$ denotes the online learning implementations.

\def \batchalgorithmic{\begin{algorithmic}
		\State initialize all parameters to zero.
		\While{$\text{episode} \leq 100$}
		\State reset the environment.
		\While{$t \leq 70$}
		\State update the network, execute $\tilde{a}_t$, and get $r_t$.
		\State append the network state, $\tilde{a}_t$, and $r_t$ to $\tau$.
		\EndWhile
		\If{every 8 episodes{\small, i.e., $(\text{episode} \% 8) = 7$}}
		\State update the parameters (e.g., with Eq.~\ref{eq:agol}).
		\State empty $\tau$.
		\EndIf
		\EndWhile
\end{algorithmic}}

\def \onlinealgorithmic{	\begin{algorithmic}
		\State initialize all parameters to zero.
		\While{$\text{episode} \leq 100$}
		\State reset the environment.
		\While{$t \leq 70$}
		\State update the network, execute $\tilde{a}_t$, and get $r_t$.
		\State append the network state, $\tilde{a}_t$, and $r_t$ to $\tau$.
		\EndWhile
		\State \textbf{if} $\tau$ size $> 70 \cdot 8$ \textbf{then} remove $\tau_0$ -- $\tau_{70}$.
		\State update the parameters (e.g., with Eq.~\ref{eq:agol}).
		\EndWhile
\end{algorithmic}}

\begin{algorithm}
	\caption{Batch learning implementation}\label{alg:batch}
	\batchalgorithmic
\end{algorithm}

\begin{algorithm}
	\caption{Online learning implementation}\label{alg:online}
	\onlinealgorithmic
\end{algorithm}

\def \simtrainingprcess{In total, this simulation experiment consisted of 20 testing conditions. The hyperparameters of each testing condition were obtained from a grid search, and the testing was repeated 10 times. Each repetition took 100 episodes with 70 timesteps per episode\review{, and the update uses the data from 8 previous episodes/runs.} To reduce the randomness for the comparison, the robot state and the neural control state were reset every episode.}

\simtrainingprcess{} The reward and advantage estimate were computed from:
\begin{equation}
A_t = \left( R_t - \mu_{R_t} \right)/ \sigma_{R_t},
\label{eq:advantagesim}
\end{equation}
\begin{equation}
R_t = \sum_{t'=t}^{70} r_{t'},
\label{eq:returnsim}
\end{equation}
\begin{equation}
r_t = (x[t]-x[\mathbin{{t}{-}{1}}]) - (y[t]-y[\mathbin{{t}{-}{1}}]),
\label{eq:rewardsim}
\end{equation}
where $A_t$, $R_t$, and $r_t$ denote the advantage, return, and reward at timestep $t$ in trajectory $\tau$, respectively, $\mu_{R_t}$ and $\sigma_{R_t}$  denote the average and standard deviation computed from all the returns at timestep $t$, and $x[t]$ and $y[t]$ denote the robot position along the x-axis (forward/backward direction) and y-axis (sideways direction) at timestep $t$ in trajectory $\tau$, as shown in Fig.~\ref{fig:morf}.

\begin{table}[!h]
	\caption{Final rewards and the number of episodes before receiving an episodic reward above 0.2 (i.e., the reward value achievable within the first 100 episodes by half of the methods tested in this work). It should be noted that the highest final reward and the lowest number of episodes before receiving an episodic reward above 0.2 are highlighted in bold. \review{The significant levels of all comparison pairs are depicted in Fig.~\ref{fig:pval}.}}
	\label{tab:rewardsummary}
	\footnotesize
	\centering
	\begin{tabular}{l  c  c  c c}
		\bottomrule
		\multirow{2.5}{*}{\makecell{\bf{\review{L}earning}\\\bf{algorithm}}} & \multicolumn{2}{c}{\bf{\review{F}inal reward}} & \multicolumn{2}{c}{\bf{\# episodes to reach 0.2}}  \\ \cmidrule(lr){2-3} \cmidrule(lr){4-5}
		& \bf{CPGRBF} & \bf{\nc} &   \bf{CPGRBF} & \bf{\nc}  \\
		\bottomrule
		\review{\bf{PG}} & -0.01 & -0.02 & \na & \na \\
		\review{\bf{PG\plus}} & -0.02 & 0.08 & \na & \na \\
		\review{\bf{PPO}} & 0.01 & 0.06 & \na & \na \\
		\review{\bf{PPO\plus}} & -0.04 & 0.0 & \na & \na\\
		\review{\bf{PIBB}} & 0.23 & 0.2 & 60 & 79\\
		\review{\bf{PIBB\plus}} & 0.34 & 0.29 & 49 & 60\\
		\review{\bf{PGPE}} & 0.31 & 0.33 & 71 & 64\\
		\review{\bf{PGPE\plus}} & 0.32 & 0.42 & 54 & 40\\ 
		\review{\bf{AGL}} & 0.36 & 0.43 & 66 & 52\\
		\review{\bf{\la}} & 0.35 & \bf{0.58} & 54 & \bf{35}\\
		\toprule
	\end{tabular}
\end{table}

\def \pvalcaption{\review{Mann-Whitney U test p-value corresponding to all comparison pairs in terms of (a) the final episodic reward and (b) the number of episodes taken to reach the reward of 0.2, where each cell intensity represents the significant level. The white cells indicate the pairs with p-value $\geq$ 0.05 (insignificant comparison), the light green cells indicate the pairs with low p-value, i.e., p-value $<$ 0.05 (significant comparison), and the dark green cells indicate the pairs with very low p-value, i.e., p-value $<< 0.05$ (highly significant comparison).}}
\begin{figure}[!h]
	\centering
	\begin{subfigure}[b]{0.37\linewidth}
		\includegraphics[width=\linewidth]{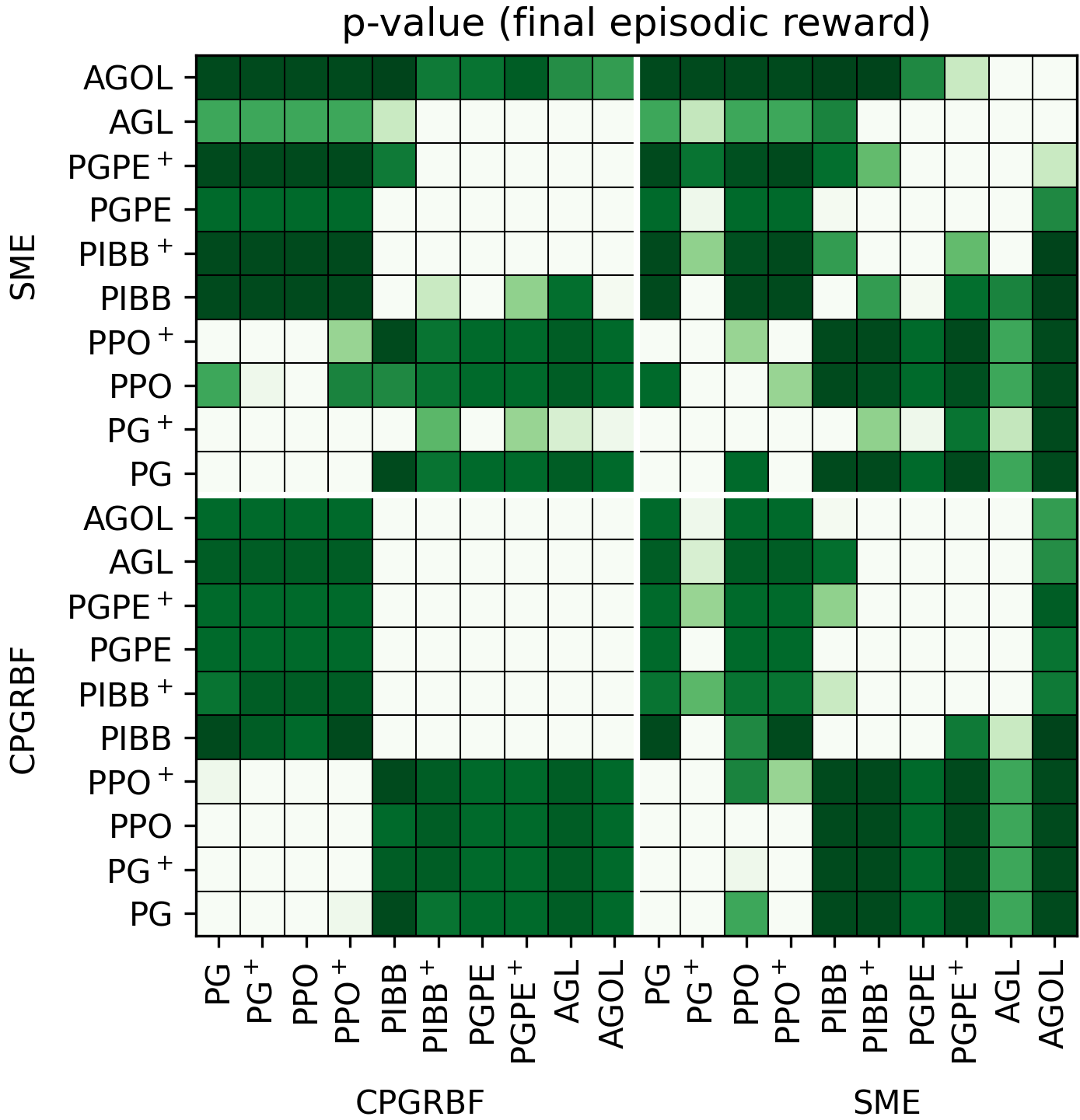}
		\caption{}
		\label{fig:preward}
	\end{subfigure}
	\begin{subfigure}[b]{0.37\linewidth}
		\includegraphics[width=\linewidth]{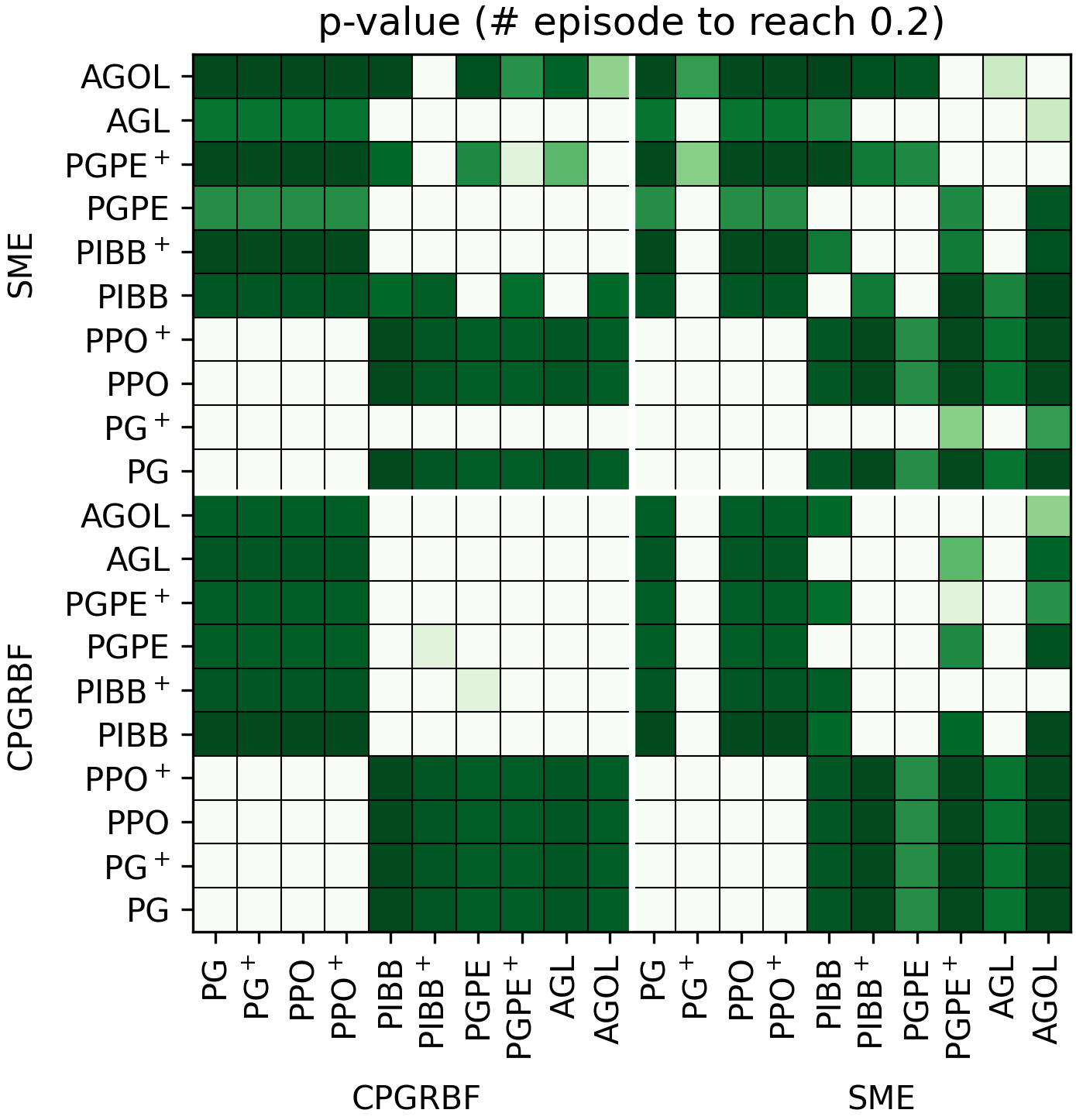}
		\caption{}
		\label{fig:pconver}
	\end{subfigure}
	\caption{\pvalcaption}%
	\label{fig:pval}
\end{figure}

\def \whyzeropointtwo{
\review{In this experiment, the comparison was performed using two matrices. The first is the final episodic reward, indicating the performance. The second is the number of episodes taken to obtain the same episodic reward value, indicating the learning speed. In this experiment, an episodic reward threshold of 0.2 was selected (i.e., a robot forward speed of 0.2 m/episode or $\approx$ \SI{20}{cm}/gait cycle) This threshold is equivalent to \SI{40}{\percent} of the theoretical maximum of the robot walking speed ($V_\text{max}$), which is computed from $V_\text{max} = 2L_\text{max}$, where $L_\text{max}$ is the maximum leg length of the robot (here, 25 cm), and a forward speed of $\approx$ \SI{60}{\%} of its body length per gait cycle (here, 30 cm). The results of this experiment are summarized in Table~\ref{tab:rewardsummary} and presented in details in \figrewardbatchcpg--\figrewardonlinesme. Besides, a video of this simulation experiment can be seen at \videolink.}}

\whyzeropointtwo



\begin{figure}[!h]
	\centering
	\begin{subfigure}[b]{0.47\linewidth}
		\includegraphics[width=\linewidth]{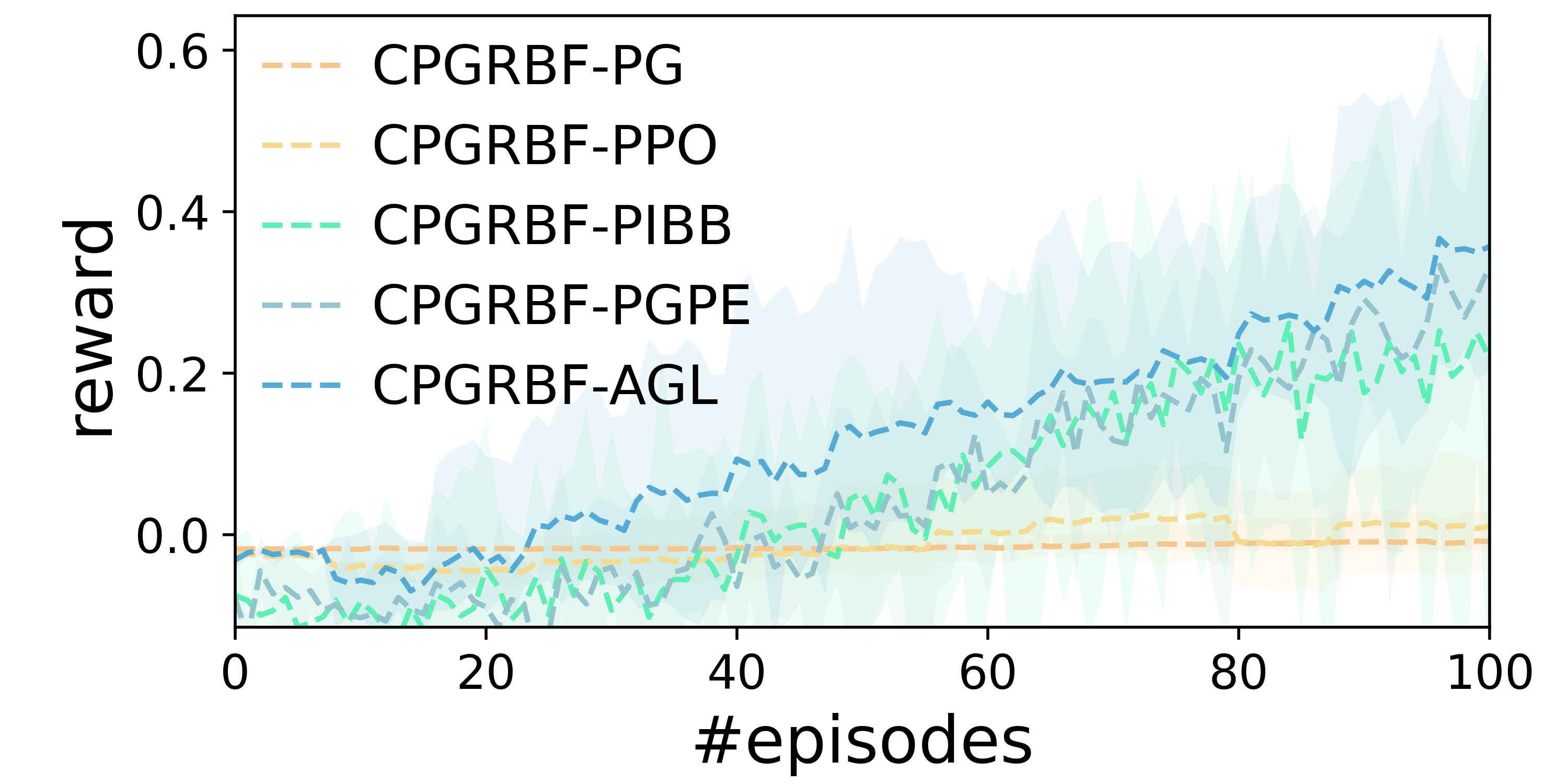}
		\caption{}
		\label{fig:cpgbatch}
	\end{subfigure}
	\hfill
	\begin{subfigure}[b]{0.47\linewidth}
		\includegraphics[width=\linewidth]{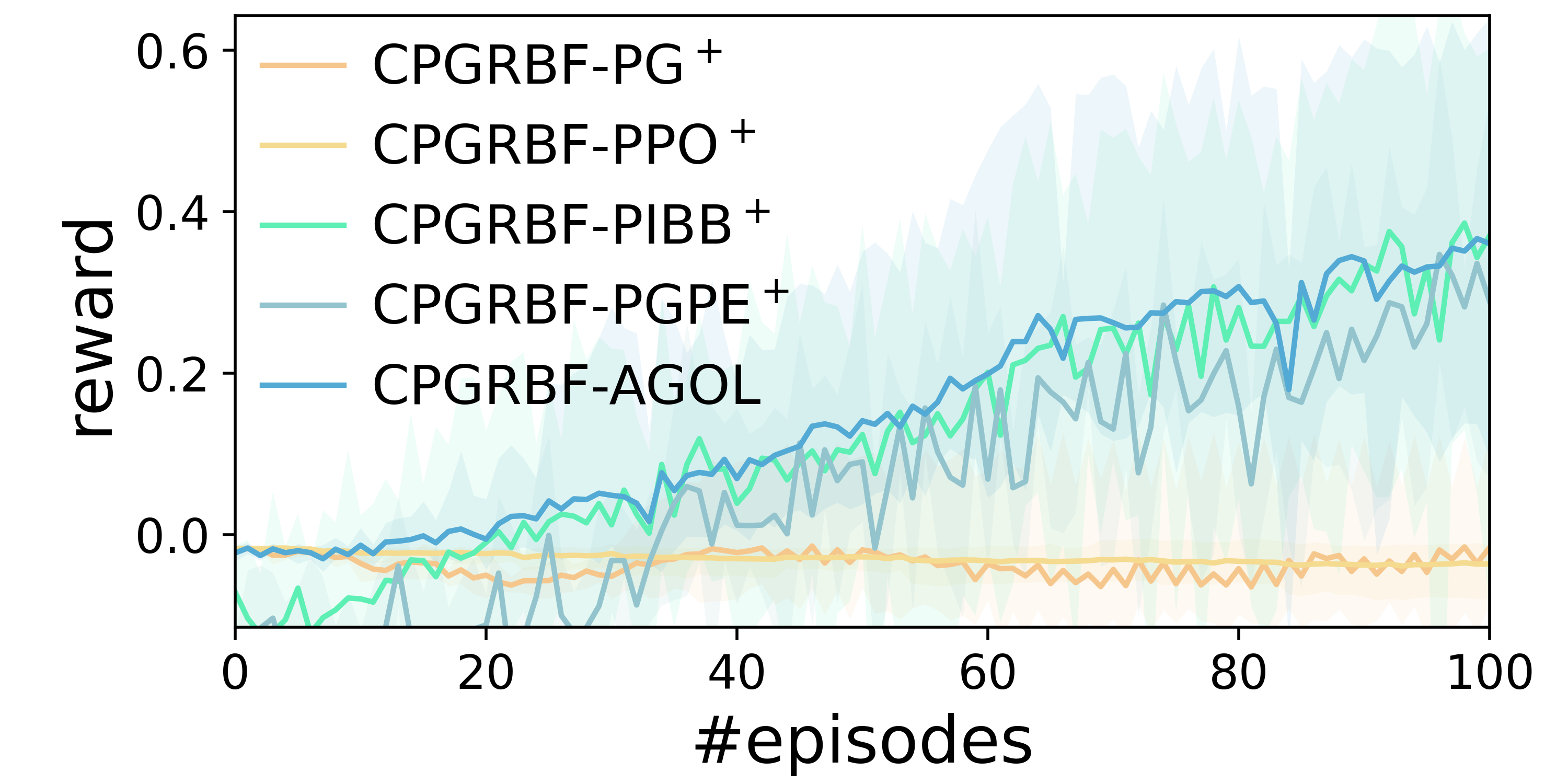}
		\caption{}
		\label{fig:cpgonline}
	\end{subfigure}
	\caption
	{\figcurve{CPGRBF}{\cite{mathias_cpgrbf}}}%
	\label{fig:cpg}
\end{figure}

\def \comparecpgrbf{
Considering first the CPGRBF neural control architecture, the combination of CPGRBF-AGL received the highest average final reward of around 0.36, reaching 0.2 after 66 episodes. The second-best performance was obtained from the one with CPGRBF-\la{} (online learning), which received a slightly lower average final reward of 0.35, reaching a reward of 0.2 after 54 episodes. Two groups of performances can be observed in both \figrewardbatchcpg{} and \figrewardonlinecpg{} \review{(p-value $<$ 0.05, Kruskal-Wallis test)}. In the first group, using CPGRBF along with parameter space learning algorithms (PIBB, PIBB\plus, PGPE, PGPE\plus, AGL, and \la) yielded average final rewards above 0.2 with no significant difference within the group \review{(p-value $>$ 0.05, Kruskal-Wallis test)}. In the second group, using CPGRBF along with all action space learning algorithms (PG, PG\plus, PPO, and PPO\plus) failed to produce a reward higher than 0.2 with no significant difference within the group \review{(p-value $>$ 0.05, Kruskal-Wallis test)}.
}

\comparecpgrbf

\begin{figure}[!h]
	\centering
	\begin{subfigure}[b]{0.47\linewidth}
		\includegraphics[width=\linewidth]{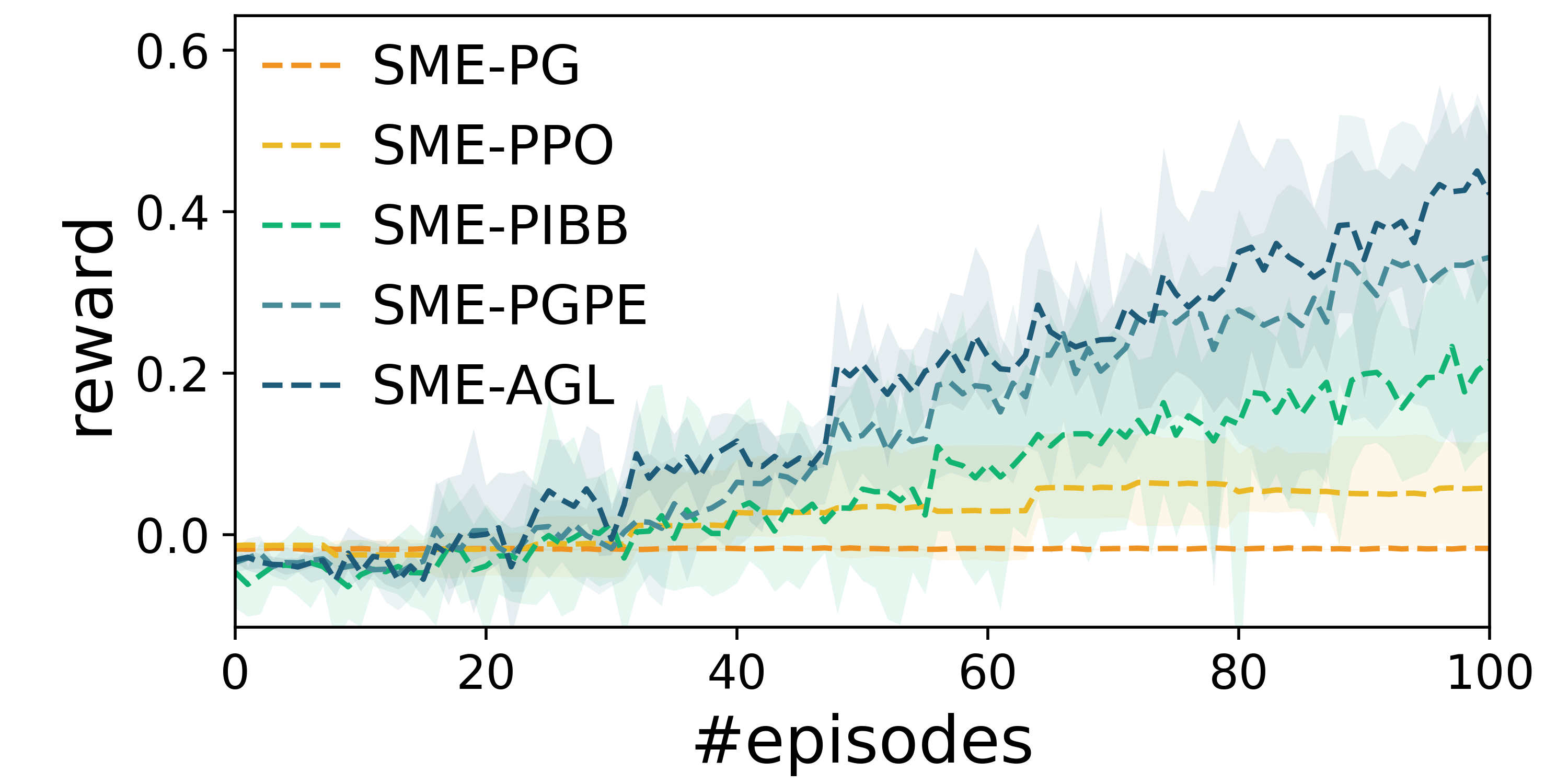}
		\caption{}
		\label{fig:smebatch}
	\end{subfigure}
	\hfill
	\begin{subfigure}[b]{0.47\linewidth}
		\includegraphics[width=\linewidth]{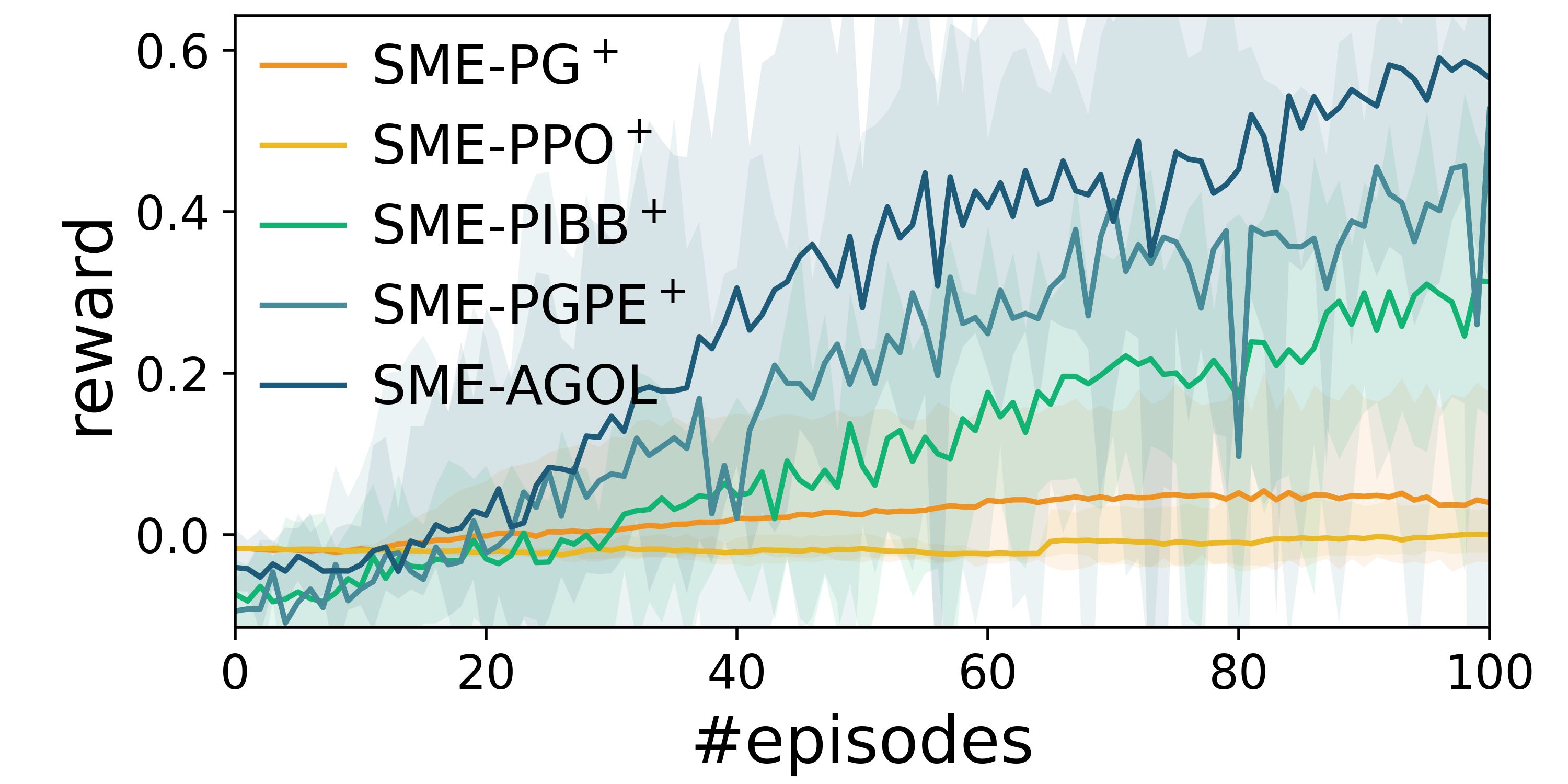}
		\caption{}
		\label{fig:smeonline}
	\end{subfigure}
	\caption{\figcurve{proposed \nc{}}{}}%
	\label{fig:sme}
\end{figure}

\def \comparesme{
Moving to the \nc{}, the combination of \nc-\la{} (the proposed method) received the highest average final reward of around 0.58 (\review{\SI{60}{\percent}} greater than the one from \nc-AGL, \review{p-value $<$ 0.05, Mann-Whitney U test}), where the reward reached 0.2 after 35 episodes. The second-best performance is from the one with \nc-AGL, which received an average final reward of 0.43, reaching a reward of 0.2 after 52 episodes. While \nc-\la{} and \nc-AGL remarkably outperformed the other combinations \review{(p-value $<$ 0.05, Mann-Whitney U test)}, there were no significant differences between the performances of \nc-\la{} and \nc-AGL \review{(p-value $>$ 0.05, Mann-Wallis test)}. Two groups of performances can be observed in Fig.~\ref{fig:smebatch} and Fig.~\ref{fig:smeonline} in a similar way to the previous results; however, the observable difference between the batch parameter space exploration learning and online parameter space exploration learning was presented \review{(p-value $<$ 0.05, Kruskal-Wallis test)}. Batch learning (\nc-PIBB, \nc-PGPE, and \nc-AGL) received the reward values between 0.2 and 0.43, while online learning (\nc-PIBB\plus, \nc-PGPE\plus, and \nc-\la{}) received between 0.29--0.58, or rather \review{\SI{15}{\percent}} more.
}

\comparesme

\def \comparenn{
Comparing the neural control architectures, i.e., between Fig.~\ref{fig:cpg} and Fig.~\ref{fig:sme}, the proposed \nc{} neural control generally outperforms the state-of-the-art CPGRBF neural control by \review{\SI{30}{\percent}} in parameter space learning, where the average final rewards are all beyond 0.2 \review{(p-value $<$ 0.05, Kruskal-Wallis test)}. \nc{} might not be suitable for implementation with PIBB, PIBB\plus{}, PGPE, and PGPE\plus{} since they yield a insignificantly different results compared to those with CPGRBFN \review{(p-value $>$ 0.05, Mann-Whitney U test)}. However, it achieves significant improvements of \review{\SI{65}{\percent}} with \la{} \review{(p-value $<$ 0.05, Mann-Whitney U test)}. It should also be noted that \nc-\la{} achieves an average final reward of more than \review{\SI{150}{\percent}} and uses \review{\SI{40}{\percent}} less time/sample to achieve a reward of 0.2 than the CPGRBF-PIBB, employed in \cite{mathias_cpgrbf} \review{(p-value $<$ 0.05, Mann-Whitney U test)}.
}

\comparenn


\begin{figure}[!h]
	\centering
	\includegraphics[width=0.8\linewidth]{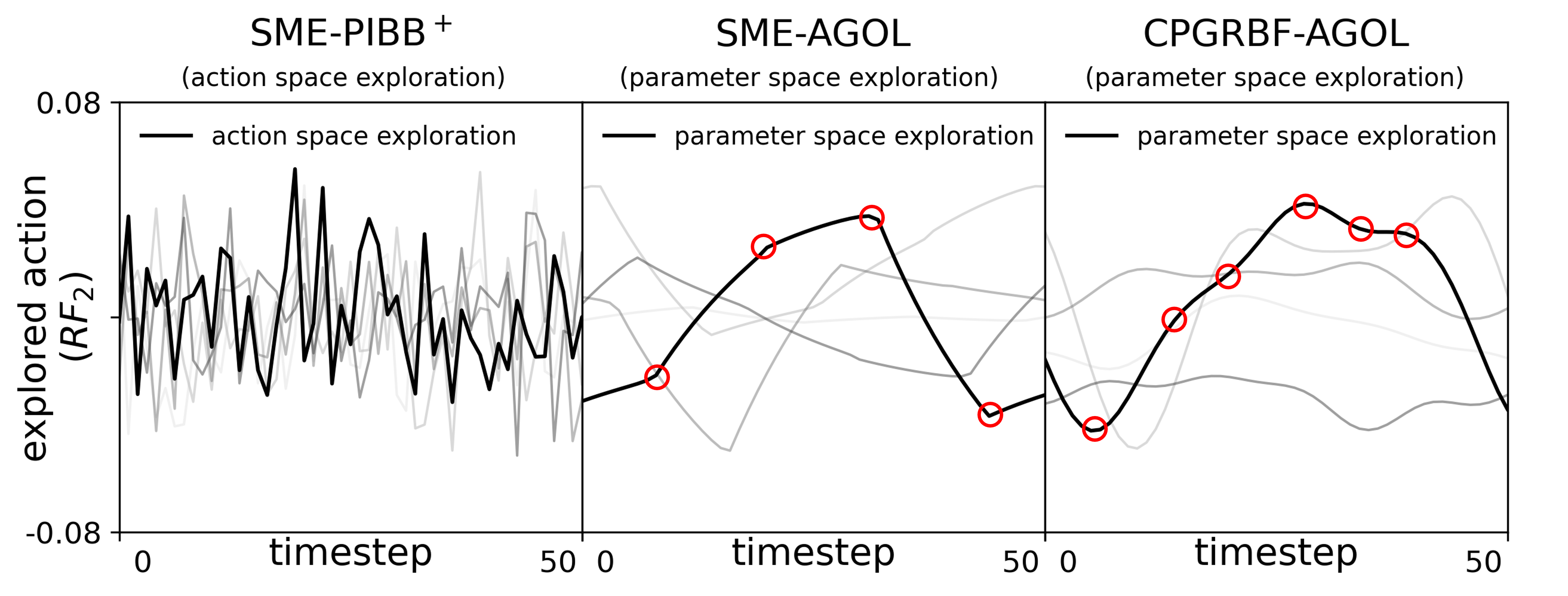}
	\caption{Five examples of the explored robot joint trajectories during the first learning episodes obtained from (left) an action space exploration learning algorithm (\nc-PIBB\plus), (middle) a parameter space exploration learning algorithm implemented with \nc{} (\nc-\la), and (right) a parameter space exploration learning algorithm implemented with CPGRBF neural control (CPGRBF-\la). It should be noted that the estimated key poses (i.e., where the trajectory slope changes significantly) are highlighted by the red circles.}%
	\label{fig:explored_action}
\end{figure}

\def \keypointp{whereas CPGRBF-\la{} learns to bend the curve, equivalent to modifying 4--7 key points \review{(p-value $<$ 0.05, Mann-Whitney U test)}}

To analyze the results, Fig.~\ref{fig:explored_action} plots different explored trajectories obtained from action space exploration and parameter space exploration. The former approach attempts different noisy joint trajectories, which causes the learning algorithms like PG and PPO to struggle in computing the updated direction, especially when using fewer samples from eight previous episodes, receiving a lower reward as a consequence. In contrast, the latter approach attempts relatively more consistent explorations: \nc-\la{} learns to modify four key points, \keypointp, as shown in Fig.~\ref{fig:explored_action}. Accordingly, parameter space exploration outperforms action space exploration, while \nc{} outperforms the CPGRBF under most conditions.

\begin{figure}[!h]
	\centering
	\includegraphics[width=0.8\linewidth]{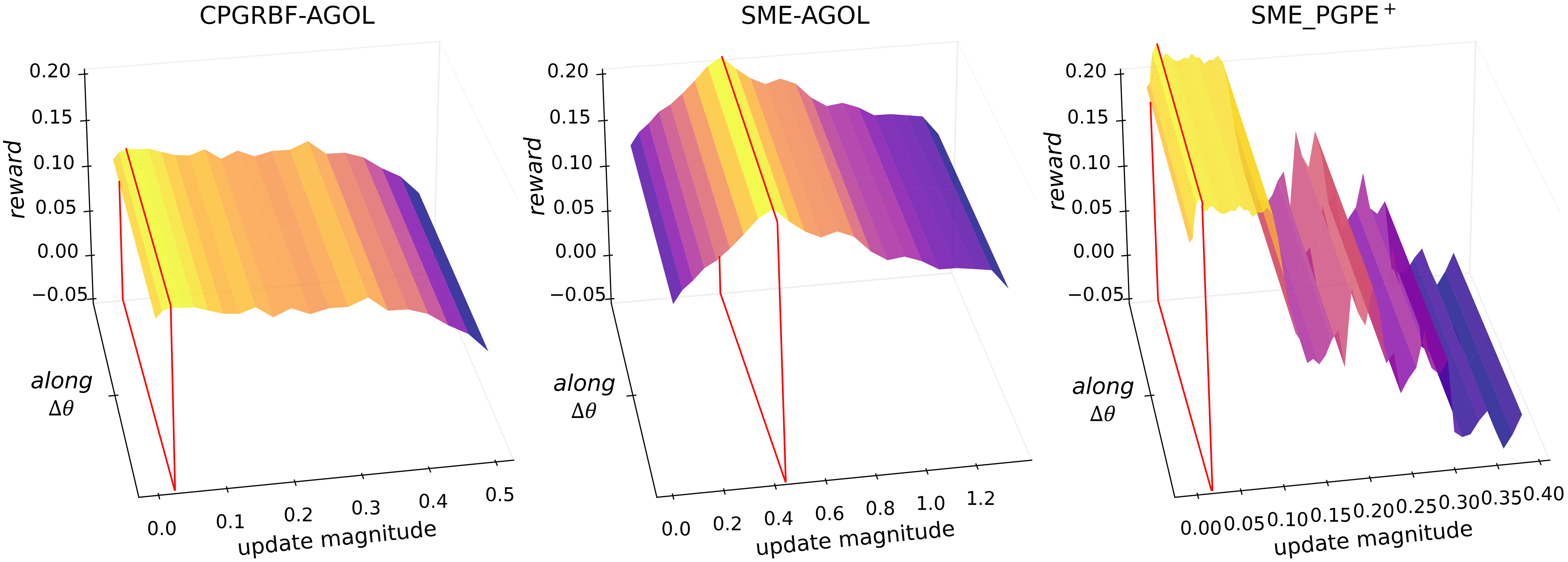}
	\caption{Reward landscape \cite{rewardlandscape} estimated along the update direction $\Delta \theta$ of (left) CPGRBF-\la, (middle) \nc-\la, and (right) \nc-PGPE\plus, where the parameter $\theta = \mbf{W}^o_b$.}%
	\label{fig:reward_landscape}
\end{figure}

\def\pgradient{In comparison, \nc-\la{} (middle), which has the correction for non-neighbor basis interference and less relevant parameters, yields a more accurate reward update estimation with comparatively fewer bumps/local maximums and an optimal step of slightly above 0.4 ($\approx$ 10 times higher, \review{p-value $<$ 0.05, Mann-Whitney U test}). }

To further analyze the accuracy of gradient estimation, Fig.~\ref{fig:reward_landscape} plots three reward landscapes, estimated along the three update directions obtained from CPGRBF-\la, \nc-\la, and \nc-PGPE\plus{} after 50 episodes. This visualization reveals that CPGRBF-\la{} (left) and \nc-PGPE\plus{} (right) have an optimal update magnitude of around 0.03; besides, the landscapes have multiple bumps/local maximums. This is due to high interference between radial bases in the former approach (Fig.~\ref{fig:basis_vs_rbf}) and interference from less relevant parameters in the latter approach (Section~\ref{sec:learning}). \pgradient

\subsection{Physical Robot Experiment}
\label{sec:expreal}

Extending from the simulation experiment, the aim of the physical experiment was to demonstrate locomotion learning with \nc-\la{} and adaptive exploration (Eq.~\ref{eq:agol_sigma}) on a physical hexapod robot. The locomotion learning with the physical robot and \nc-\la{} was repeated 10 times, \review{following Algorithm~\ref{alg:real}} with the empirically chosen learning rate of $\eta_\theta = 0.5$ and $\eta_\sigma = 0.1$. Due to the space limit of the experimental area, the experiment was paused when the robot reached the end of the testing platform and continued when moving it manually to the starting location. Each repetition took 200 episodes with 70 timesteps per episode, \review{the data from 8 previous episodes}, and without robot state/network state reset. 

\def \realalgorithmic{
	\begin{algorithmic}
		\State initialize all parameters to zero.
		\While{true}
		\While{$t \leq 70$ ($\approx$ 1 gait cycle)}
		\State update the network (Eqs.~\ref{eq:zpg}--\ref{eq:output}) and execute $\tilde{a}_t$.
		\State compute $r_t$ according to Eq.~\ref{eq:rewardreal}.
		\State append the network state, $\tilde{a}_t$, and $r_t$ to $\tau$.
		\EndWhile
		\State \textbf{if} $\tau$ size $> 70 \cdot 8$ \textbf{then} remove $\tau_0$ -- $\tau_{70}$.
		\State compute advantage according to Eqs.~\ref{eq:advantagereal}.
		\State update the parameters with Eqs.~\ref{eq:agol} and \ref{eq:agol_sigma}.
		\State update the baseline estimation by minimizing Eq.~\ref{eq:bloss}. 
		\EndWhile
	\end{algorithmic}
}

\begin{algorithm}
	\caption{\nc-\la{} implementation (physical robot)}\label{alg:real}
	\realalgorithmic
\end{algorithm}

\def \realprocessdescribe{
\review{All parameters were initialized as zero before three iterative steps were performed. This process was updated until the end of the experiment. First, the robot performed exploration for 70 timesteps ($\approx$ 1 gait cycle), by executing explored actions ($\tilde{a}_t$) obtained from the \nc{} network. The reward ($r_t$) was then computed according to:} 
\begin{equation}
r_t = \Delta x[t] \cos{(\Delta \psi[t])} + \Delta y[t] \sin{(\Delta \psi[t])}, \lab{eq:rewardreal}
\end{equation}
where $r_t$ denote the reward at timestep $t$, while $\Delta x[t]$, $\Delta y[t]$, and $\Delta \psi[t]$, respectively denote the robot displacement at timestep $t$ in x direction, y direction, and yaw \review{orientation}. In this work, $\Delta x[t]$ $=$ $x[t]-x[\mathbin{{t}{-}{1}}]$,  $\Delta y[t]$ $=$ $y[t]$ $-$ $y[\mathbin{{t}{-}{1}}]$, and $\Delta \psi[t]$ $=$ $\psi[t]-\psi[\mathbin{{t}{-}{1}}]$, where the estimated robot pose ($x[t]$, $y[t]$, and $\psi[t]$) are obtained from the tracking camera, as shown in Fig.~\ref{fig:morf}. \review{Subsequently, the network states (i.e., activities and parameters), explored actions, and rewards were appended to the trajectories/training samples.}

\review{Second, using the training samples from the previous 8 episodes, the advantage values were estimated according to:}
\begin{equation}
A_{t} = \left( R_t - \mbf{W}^v_b \mbf{b}[t] \right)/ \text{RMS}_\tau(R_t - \mbf{W}^v_b \mbf{b}[t]),
\lab{eq:advantagereal}
\end{equation}
\begin{equation}
R_t = \sum_{t'=t}^{70} r_{t'},
\lab{eq:returnreal}
\end{equation}
where $A_t$, $R_t$, and $r_t$ denote the advantage, return, and reward at timestep $t$ in the trajectory $\tau$, respectively, $\mbf{W}^v_b \mbf{b}[t]$ denotes the baseline, $\text{RMS}_\tau()$ denotes the root-mean-square function computed over the trajectory $\tau$, $\mbf{b}[t]$ denotes the activity vector of the bases ($b_i[t]$) at timestep $t$ in trajectory $\tau$, and $\mbf{W}^v_b$ the mapping weights from Bs to baseline prediction. Given that this advantage estimation incorporates the predicted baseline/expected value at each specific state, the robot learning can operate without the robot state reset, which is practical for physical robot learning.

\review{Third, the policy parameters (i.e., action generation-related parameters) were updated according to the learning rule in Eq.~\ref{eq:agol} and the exploration parameters were updated according to the learning rule in Eq.~\ref{eq:agol_sigma}. This was then followed by updating the baseline prediction, which was performed using gradient descent with a learning rate of 0.05 to minimize the mean square error loss:}
\begin{equation}
\mathcal{L} = \text{MSE}_\tau(R_t- \mbf{W}^v_b \mbf{b}[t]),
\lab{eq:bloss}
\end{equation}
\review{where $\mathcal{L}$ denotes the loss function and $ \text{MSE}_\tau()$ denotes the mean square error between the actual return $R_t$ and the baseline prediction $\mbf{W}^v_b \mbf{b}[t]$, averaged over a trajectory $\tau$.}
}

\realprocessdescribe

\begin{figure}[!h]
	\centering
	\begin{subfigure}[b]{0.6\linewidth}
		\includegraphics[width=\linewidth]{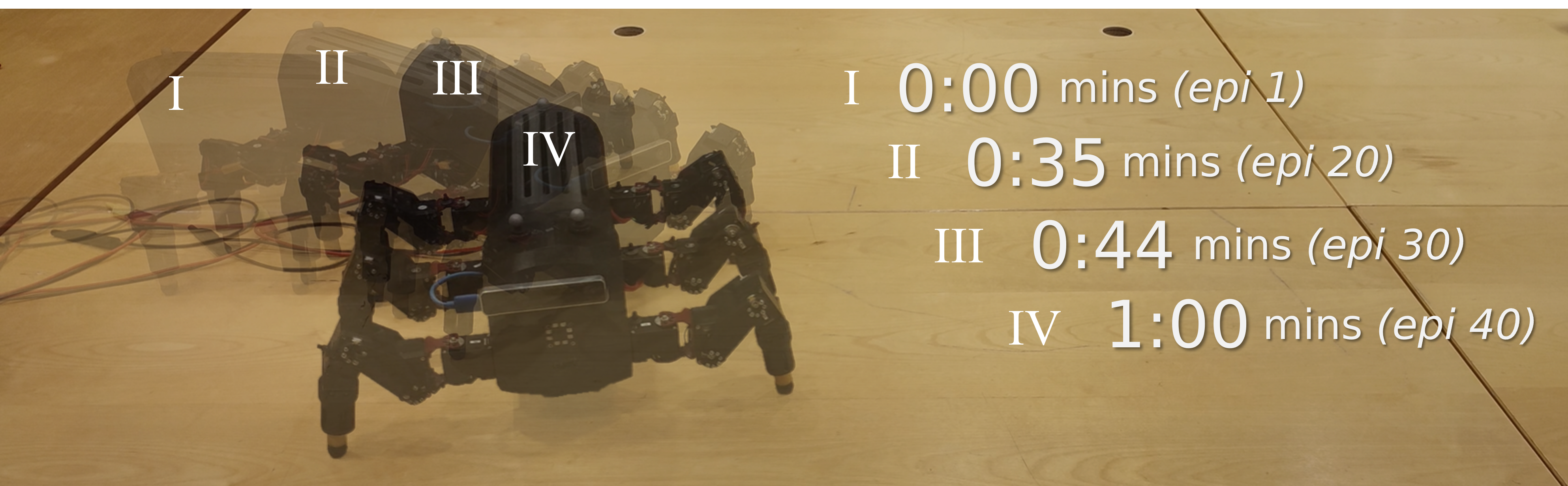}
		\caption{}
		\label{fig:realsnap1}
	\end{subfigure}
	\\
	\centering
	\begin{subfigure}[b]{0.6\linewidth}
		\includegraphics[width=\linewidth]{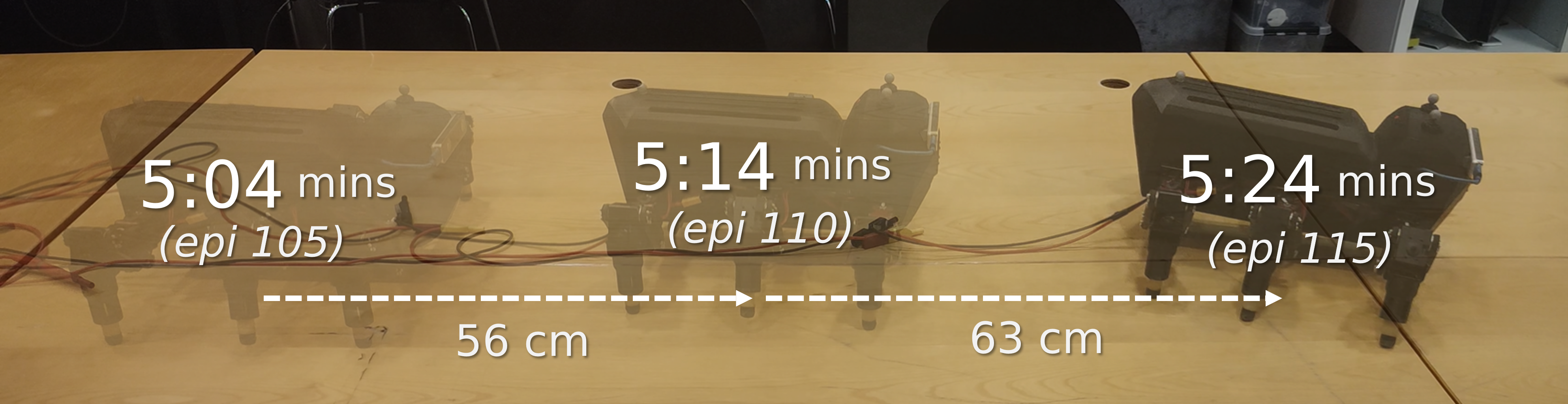}
		\caption{}
		\label{fig:realsnap2}
	\end{subfigure}
	\caption{Snapshots from a trial of physical robot locomotion learning (a) during the first 40 episodes and (b) between 105 episodes and 115 episodes. The video of the experiment is available at \videolink.}%
	\label{fig:realsnap}
\end{figure}

\begin{figure}[!h]
	\centering
	\includegraphics[width=0.8\linewidth]{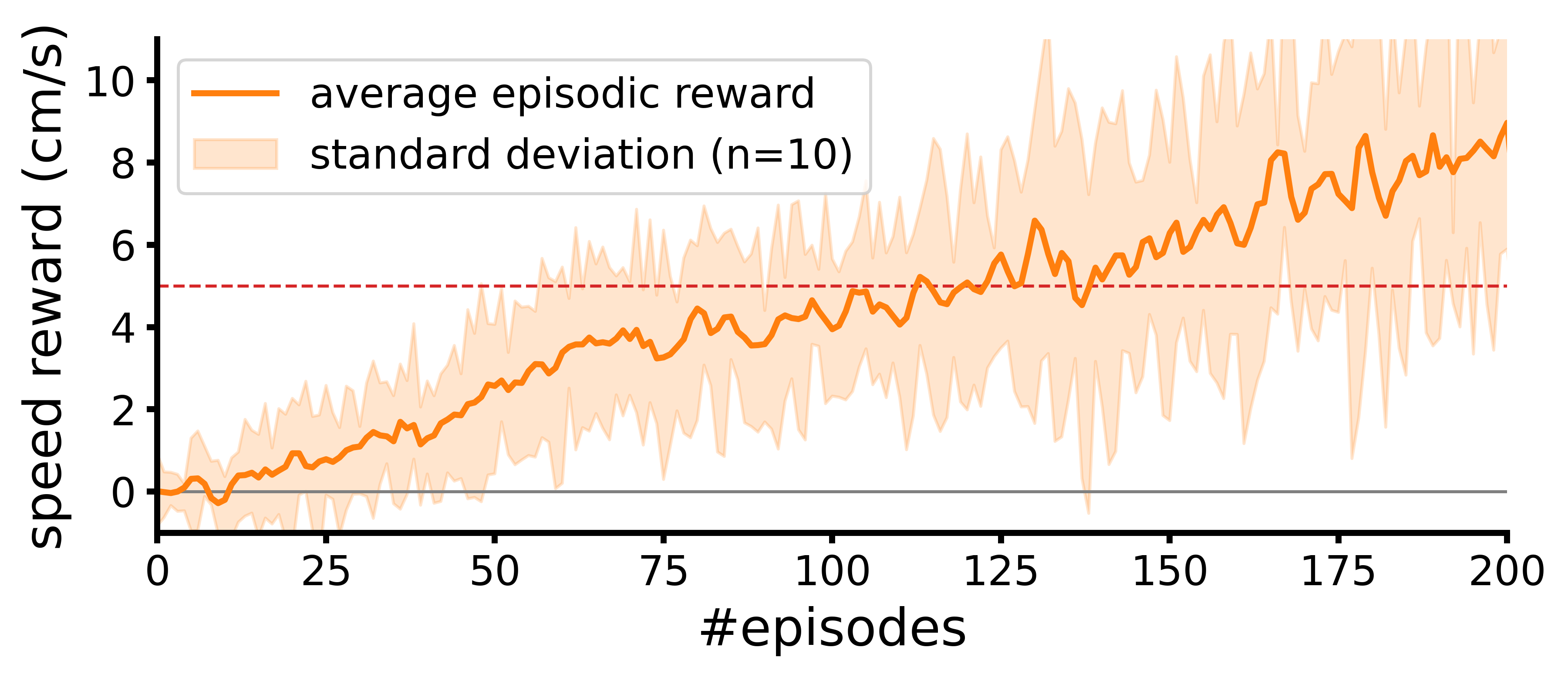}
	\caption{Average episodic reward obtained from 10 repetitions of physical robot locomotion learning, along with the standard deviation. The red dashed line indicates a locomotion speed of 5 cm/s, which is also demonstrated by the same physical robot with a manually designed locomotion controller presented in \cite{pbird_morfendocrine}.}%
	\label{fig:realreward}
\end{figure}

The locomotion learning result is presented in Fig.~\ref{fig:realsnap}, and a supplementary video can be viewed at \videolink, while the corresponding speed (i.e., computed from the reward) is provided in Fig.~\ref{fig:realreward}.

Fig.~\ref{fig:realreward} presents that the average speed starts around 0.0 cm/s, indicating that locomotion learning begins without any pre-trained knowledge. The robot then takes merely 20 episodes of trial-and-error ($\approx$ 35 s) before it starts moving forward, receiving a speed of nearly 1.0 cm/s, as also shown in \figresnapa. During the first 40 episodes ($\approx$ 1 min), the robot exhibited a small positive speed of almost 2.0 cm/s while turning to the right, as shown in \figresnapa, due to the poor coordination between left and right legs. After 50 subsequent episodes ($\approx$ 2 mins), the robot's curved locomotion path was corrected, and it eventually developed almost straight locomotion with a speed of around 5 cm/s. Additionally, \figresnapb{} illustrates that the robot walked for 56 cm in 10 s between the 105$^{\text{th}}$ episode and 110$^{\text{th}}$ episode, equivalent to the one measured from the same physical robot with a manually designed locomotion controller demonstrated in \cite{pbird_morfendocrine} (i.e., $\approx$ 5 cm/s). The robot then maintained straight locomotion and continued improving its speed, resulting in an improved travel speed of 63 cm in 10 s around 115 episodes, as illustrated in \figresnapb. By 200 episodes, the robot achieved an average speed of slightly above 8 cm/s (between 6 cm/s to 11 cm/s in 10 experiment trials). 

\begin{figure}[!h]
	\centering
	\includegraphics[width=0.95\linewidth]{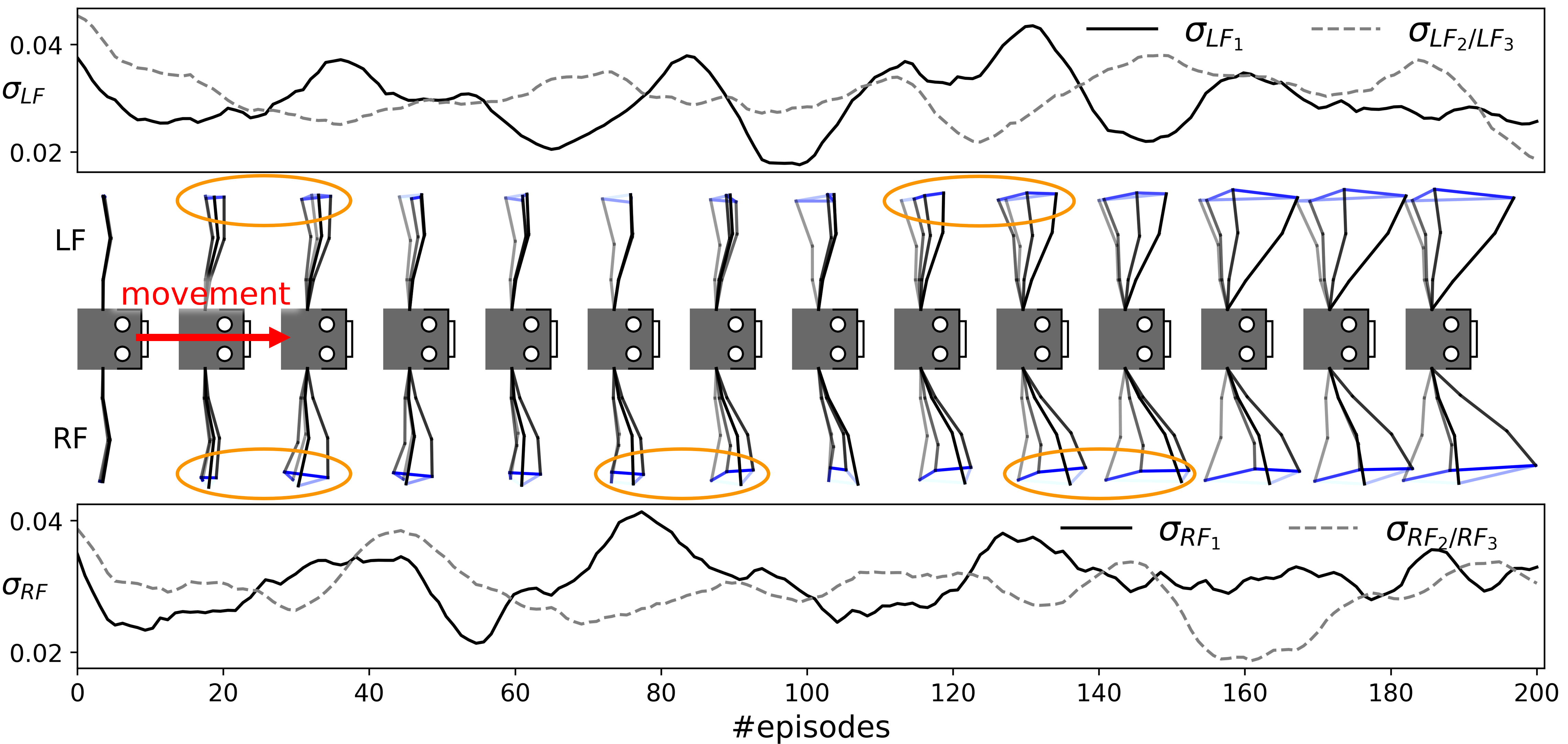}
	\caption{Graphical representation of the top view robot kinematics and evolution of the adaptive exploration rates of the left front (LF) and right front (RF) legs obtained from the locomotion learning in Fig.~\ref{fig:realsnap}. In the graphical representation, four key poses are represented by four gray shades: black $\rightarrow$ dark gray $\rightarrow$ gray $\rightarrow$ light gray $\rightarrow$ back to black, with the corresponding foot heights represented by blue shades: dark blue indicating a relatively lower foot path (stance phase), and light blue a relatively higher foot path (swing phase). For visualization purposes, small offsets (i.e., 0.1 rads) are added to all joints.}%
	\label{fig:realsigma}
\end{figure}

In order to analyze the physical robot locomotion learning, Fig.~\ref{fig:realsigma} plots the evolution of leg kinematics and adaptive exploration rates obtained from a locomotion learning trial. The leg kinematics is visualized using the output mapping connection weights $\mbf{W}^o_b$ (18 motor positions/rows $\times$ four key poses/columns) along with the kinematics model, while the adaptive exploration rates of the swing joints (e.g., $\text{RF}_1$) and leg lifting joints (e.g., $\text{RF}_2$ and $\text{RF}_2$) plotted directly from the moving averages of $\sigma_{LF_1}$ and $\sigma_{RF_1}$ and those $\sigma_{LF_2/LF_3}$ and $\sigma_{RF_2/RF_3}$, receptively.

Fig.~\ref{fig:realsigma} reveals that the robot autonomously varies its exploration rate, and this could be used to interpret the focus of the learning. For instance, the overlapping exploration rates between swing joints (e.g., $\text{RF}_1$, black lines) and leg lifting joints (e.g., $\text{RF}_2$ and $\text{RF}_3$, gray dashed lines) indicate that the robot repeatedly shifts its focus between learning the lifting actions and swinging actions. In the first 30 episodes, greater exploration of leg lifting joints (gray dashed lines above black lines) suggests a focus on adjusting the lifting pattern. After obtaining a proper lifting pattern (e.g., after 30 episodes, as also shown in Fig.~\ref{fig:realsnap}), focus shifted toward the swing/shoulder joints (black lines above gray dash) to increase the swing amplitude. This process is repeated for around 30, 80, and 130 episodes, while an increase in swing amplitudes can be observed from the leg kinematics for most of the time due to random exploration, as highlighted in orange.

\section{Discussion and Conclusion} \label{sec:conclusion}

To deal with sample inefficiency and the interpretability of reinforcement learning-based robot locomotion learning, this study verifies that incorporating the neural architecture and learning algorithm with interpretability can also enhance sample efficiency and overall performance.

\def \interpconclude{\review{Considering the \nc{} neural control, its interpretable network structure, neural activities, and synaptic connections are utilized. In terms of network structure, the \nc{} is equipped with \review{two interpretation dimensions}. Three layers are employed to separate the network function vertically (Fig.~\ref{fig:neuralcontrol}), providing layer-wise interpretation along with minimal learning parameters (i.e., only the last mapping layer), and four neuron columns are employed to separate neural activities horizontally (Fig.~\ref{fig:neuralcontrol}), providing less interference of the non-neighbor robot states/movement bases \cite{reviewInterpretableRL}. The network structure with last-layer learning differs from conventional fully connected end-to-end control networks (FCNN) \cite{MELA,HlifeRL_optionbased,dogrobot_massivelyparallel} and offers the following advantage. It requires fewer parameters to train, leading to faster convergence compared to full end-to-end learning (Table~\ref{tab:learningspeedsummarize}). In terms of neural activities, the generated bases exhibit less interference between distal neurons (non-neighbor neurons). This leads to \SI{33}{\percent} less interference compared to the CPGRBF neural control \cite{mathias_cpgrbf} and \SI{10}{\percent} less compared to a fully connected neural network (FCNN) with ReLU activation function (Section~\ref{sec:neuralcontrol}). Note that the FCNN, considered as a black-box control method \cite{reviewInterpretableRL}, typically produces high interference activation patterns (Section~\ref{sec:fcnnactivity} of the supplementary material); thus, around \SI{17}{\percent} of each neuron activity is interfered by at least one of the others. Therefore, the \nc{}, which employs reduced interference bases further facilitates the interpretation and collective learning \cite{RLbook}. Finally, in terms of synaptic connections, the connection weight parameters are equipped with their own interpretations/meanings. For example, the mapping weights encode locomotion pattern/shape (as shown in Section~\ref{sec:differentshape} of the supplementary material) and leg coordination (as shown in Section~\ref{sec:legcoordination} of the supplementary material), providing learning transparency and less interference update \cite{reviewInterpretableRL}. In another example, $w_\tau$ solely represents the frequency/state transition speed parameter, further reducing the training parameters while allowing us to adjust the locomotion frequency after training (as demonstrated in \videolink{} at 2:40 mins). In terms of learning performance, the \nc{} improves the accuracy of estimated gradients and smoothens the estimated reward landscape, leading to \review{\SI{65}{\percent}} more reward and requiring \review{\SI{35}{\percent}} fewer samples in comparison to  \nc-\la{} and CPGRBF-\la{} (Section~\ref{sec:neuralcontrol}).}
}

\interpconclude



Considering the learning algorithm, the proposed \la{} rule consistently exploits parameter spaces and provides explanations \cite{lrp,gradcam} to the robot for scaling the update, allowing it to focus on more relevant parameters. Although \cite{updateimprovewithxai} discusses various forms of gradient/relevance to adapt the learning, this study further analyzes and presents the reason for applying the absolute of gradient/relevance to adapt the parameter space exploration-based learning algorithm. The findings indicate that it can adapt the learning rule when the explored action $\tilde{a}_t$ does not depend on the given explored parameters $\tilde{\theta}_t$ ($\nabla_\theta p(\tilde{a}_t|\tilde{\theta}_t) = 0$), as discussed in Section~\ref{sec:learning}, \review{which could further exploit the separated activities of \nc{}}. Thus, \la{} is capable of accurately estimating the weight update (control parameter update) using a small sample size (eight episodes), resulting in a smoother reward landscape with a further optimal step (Fig.~\ref{fig:reward_landscape}), \review{\SI{38}{\percent}} more reward, and \review{\SI{12}{\percent}} fewer samples required in comparison to \nc-\la{} and \nc-PGPE\plus.

\begin{table}[!h]
	\caption{Comparison of the robot training platforms and the learning time in timesteps, episodes, and simulation/real time across different locomotion learning approaches evaluated on flat terrain. $\#$timesteps denotes the number of controller/system update steps, $\#$episodes denotes the number of trajectories collected, and time denotes the training time without parallel training or simulation acceleration, computed from $\#$timesteps taken until the learning converges or is reported by the original articles and the controller update rate.}
	\newcommand*{\tsbar}[1]{\colorbox{red!85}{\makebox(#1mm,5){}}~}
	\label{tab:learningspeedsummarize}
	\footnotesize
	\centering
	\begin{tabular}{c | c | c | c | l| l| l}
		\bottomrule
		\multirow{2}{*}{\textbf{Method}} & \multirow{2}{*}{\textbf{Trained Robot}} & \multirow{2}{*}{\textbf{$\#$Motors}} & \multirow{2}{*}{\textbf{$\#$Outputs}} &
		\multicolumn{3}{c}{\textbf{Training Time}}
		\\ \cline{6-7}
		 & & & & \makecell{\centering \textbf{$\#$Timesteps}} & \makecell{\centering \textbf{$\#$Episodes}} & \makecell{\centering \textbf{Time}}
		\\ \toprule \bottomrule
		
		\cite{kaise_scirobotics} & \rd{sim} quadruped & 12& 12  & \tsbar{15}400m & \tsbar{13}45k & \tsbar{10}2 days
		\\
		\cite{dogrobot_massivelyparallel} & \rd{sim} quadruped & 12& 12 & \tsbar{11}30m & \tsbar{15}600k & \tsbar{13}7 days
		\\
		\cite{dreamwaq} & \rd{sim} quadruped & 12& 12  & \tsbar{13}200m & \tsbar{3.7}1k &  \tsbar{15}46 days
		\\
		\cite{KeepLearning} & \rd{sim} quadruped & 12& 12 & \na & \tsbar{3.85}1.2k\rd{$^a$} & \tsbar{10}2 days\rd{$^a$}
		\\
		\cite{MELA} & \rd{sim} quadruped & 12 & 12 & \tsbar{8}1m & \tsbar{0.1}200 & \tsbar{8.01}10 hrs \\ 
		
		\cite{HlifeRL_optionbased} & \rd{sim} other types\rd{$^b$} & 6& 6 & \tsbar{9}1.2m & \tsbar{3.85}1.2k & \tsbar{15}10 days
		\\
		\cite{hexapodlearning_fc_sixmodule} & \rd{sim} hexapod & 18& 18  & \na & \tsbar{7.61}2.3k & \na
		\\
		\cite{DEAC} & \rd{sim} quadruped & 12& 12 & \tsbar{10}2m & \na &  \tsbar{8.5}10 hrs
		\\
		\cite{dogrobot_teacherstudent} & \rd{sim} quadruped & 12& 12 & \tsbar{15}400m & \na & \tsbar{5}3 hrs
		\\		
		\cite{viabilityGaitTransition} & \rd{sim} quadruped & 12& 12  & \tsbar{11}35m & \na &  \tsbar{11}4 days
		\\
		\cite{hexdog1} & \rd{sim} hexapod & 18& 18  & \tsbar{11}5m & \tsbar{11}5k & \tsbar{10}2 days
		\\
		\cite{spikingtnnls} & \rd{sim} other types\rd{$^b$} & 3--8& 3--8 & \tsbar{8}1m & \tsbar{3.7}1k & \na
		\\

		
		\cite{fourleg_cpg_multihead} & \rd{sim} quadruped & 12& 12 & \tsbar{2.5}200k & \tsbar{0.1}100 & \tsbar{8.01}6 hrs
		\\
		\cite{LILAC_latent_method} & \rd{sim} other types\rd{$^b$} & 3--8& 3--8  & \tsbar{2}100k & \tsbar{3.7}1k & \tsbar{2.95}1 hrs
		\\
		\cite{mathias_cpgrbf} & \rd{sim} hexapod & 18& 18 & \tsbar{3.5}300k & \tsbar{3.85}1.2k & \tsbar{3.3}1.5 hrs
		\\
		\cite{mathias_cpgrbf} & \rd{sim} hexapod & 18& 9\rd{$^c$}  & \tsbar{2.5}200k & \tsbar{3.85}1.2k & \tsbar{3.3}1.5 hrs
		\\
		\cite{mathias_cpgrbf} & \rd{sim} hexapod & 18& 3\rd{$^c$}  & \tsbar{0.5}60k & \tsbar{0.2}200 & \tsbar{0.36}20 mins
		\\
		
		\cite{hexapodlearning_cpg_spikingandforce3leg} & \rd{sim} hexapod & 12& 6\rd{$^c$} & \tsbar{0.01}66\rd{$^d$} & \na & \tsbar{0.1}5 mins\rd{$^d$}
		\\ 
		\cline{1-7}
		\multirow{3}{*}{\makecell{\nc--\\\la}} & real hexapod & 18& 18 & \tsbar{0.15}14k & \tsbar{0.44}200 & \tsbar{0.26}10 mins
		\\
		& \rd{sim} hexapod & 18& 18 & \tsbar{0.1}10k & \tsbar{0.1}100 & \tsbar{0.1}5 mins
		\\
		& \rd{sim} quadruped & 12& 12  & \tsbar{0.15}14k & \tsbar{0.44}200 & \tsbar{0.26}10 mins
		\\
		\toprule
	\end{tabular}
	\begin{tablenotes}\footnotesize
		\item[] \rd{$^a$} \review{indicates the use of guided motion.}
		\item[] \rd{$^b$} \review{indicates the use of other types of robots, e.g., hopper, half-cheetah, or bipedal robot.}
		\item[] \rd{$^c$} \review{highlights the use of output encoding where the number of network outputs is less than that of the actuators/motors.}
		\item[] \rd{$^d$} \review{indicates the use of a predefined leg pattern to learn merely interlimb coordination.}

	\end{tablenotes}
\end{table}


\def \nosimtoreal{Combining \nc{} and \la{}, locomotion learning can be achieved within 200 episodes ($\approx$ 10 mins) directly on a physical hexapod robot starting from scratch, \review{i.e., all parameters were initialized to zero before learning. This demonstrates sample efficient locomotion learning in the real world, which could provide an alternative method to sim2real transfer, eliminating the need for accelerated accurate simulations \cite{dogrobot_massivelyparallel,kaise_scirobotics} or bridging the reality gap \cite{dogrobot_teacherstudent}.}}

 \nosimtoreal{} The robot takes merely around 100 episodes ($\approx$ 5 mins) on average to obtain a forward speed of 5 cm/s, reaching the performance benchmark obtained from a manually designed controller tested on the same robot \cite{pbird_morfendocrine}. Even on a low-friction terrain like an ice rink, the robot succeeds in learning forward locomotion under a similar timescale (not shown here but supplemented at \videolink). While previous works included an action smoothness reward \cite{MELA,mathias_cpgrbf,LILAC_latent_method,dogrobot_massivelyparallel}, slippage penalty \cite{mathias_cpgrbf}, or clearance/collision penalty \cite{MELA,dogrobot_massivelyparallel} for smoothing the reward landscape, \nc-\la{} achieves locomotion learning on flat terrain and ice without any of these (or without a complex reward function). It uses only a simple speed-related reward to successfully learn robot locomotion within a few minutes.

The comparison in Table~\ref{tab:learningspeedsummarize} presents that while others require either 150,000 -- 400 million timesteps or up to 600,000 episodes of trial-and-error samples collected in simulation ($\approx$ 1 hr -- 46 days) \cite{kaise_scirobotics,dogrobot_massivelyparallel,dreamwaq}, \nc-\la{} demonstrates the use of merely 10,000 timesteps and 100 episodes in simulation ($\approx$ 5 mins). Even with hexapod robots, the best performing CPGRBF neural control architecture plus the PIBB learning algorithm \cite{mathias_cpgrbf} also took 1,200 episodes ($\approx$ 1.5 hrs) in simulation to learn independent leg patterns and 200 episodes ($\approx$ 20 mins) to learn shared diagonal leg patterns. Due to such a long training time, \nc-\la{} is among only two methods demonstrating locomotion learning on physical hexapod robots, alongside \cite{hexapodlearning_cpg_spikingandforce3leg}, which incorporates output encoding by learning/adapting merely predefined leg coordination (interlimb coordination). Therefore, \nc-\la{} is the only method, to the best of our knowledge, that demonstrates sample efficient locomotion learning on a physical robot without prior knowledge. It also allows the robot to learn the patterns of all legs independently and develop both intralimb and interlimb coordination simultaneously, utilizing arbitrary trajectories within the workspace to solve the locomotion task.


\def \generization{\review{The proposed \nc-\la{} framework demonstrates a certain degree of generalization, allowing its direct application to different legged robot learning.} The \nc-\la{} method is not limited only to the tested hexapod robot since the output dimension (i.e., output mapping matrix, $\mbf{W}^o_b$) can be modified for other types of robots, \review{as shown in \videolink{} at around 3 mins (four-legged version of MORF where two middle legs were amputated) and \videobone{} (Unitree B1 quadruped robot).}} 

\def \transferability{\review{However, its transferability across different robot platforms (learning from one type and applying to another) remains limited. While transferring the learned control policy of \nc-\la{} between hexapod robots is possible, performance degradation is expected depending on the structural differences between the two robots (e.g., leg length, joint torque, and robot dynamics). Therefore, in future work, we will investigate improving \nc-\la's transferability through techniques like fine tuning \cite{KeepLearning} or domain randomization \cite{dogrobot_teacherstudent}.}}
	

\generization{}  \transferability

Apart from these aspects, multiple \nc s with in-between connections could allow the robot to learn and switch between behaviors (i.e., sets of key poses). With this mechanism, the activities in the first set can propagate to another when certain internal state neurons in the first set are inhibited by the sensory feedback, allowing the activity to propagate to another path or the second set \cite{interpRL_dooperation}. We will investigate this in the future. More importantly, besides contributing \nc-\la{} as a method for online locomotion learning, this work demonstrates the connection between interpretation and improved performance, encouraging the use of interpretation as a mechanism for obtaining understandable results and gaining a better performance toward explainable AI (XAI)-based locomotion control.

\bibliographystyle{unsrtnat}
\bibliography{references}  

\begin{center}
	\textbf{\large Supplementary Material}
\end{center}
\setcounter{equation}{0}
\setcounter{figure}{0}
\setcounter{table}{0}
\setcounter{section}{0}
\makeatletter
\renewcommand{\theequation}{S\arabic{equation}}
\renewcommand{\thefigure}{S\arabic{figure}}
\renewcommand{\thesection}{S\arabic{section}}
\renewcommand{\thetable}{S\arabic{table}}

\section{Example of Leg Coordination Patterns}
\label{sec:legcoordination}

\def\weightsetmeaning{Note that the mapping weights are presented as \{$w_1$, $w_2$, $w_3$, $w_4$\}, where $w_i$ denotes a mapping weight from the $i^{th}$ basis to the output ($LF_1$ or $RF_1$), and the $i^{th}$ key pose is $\approx$ 0.5$\cdot w_i$.}

\def\legcozerocap{\review{(left, blue) LF1 and (right, blue) RF1 commands, presented along with their shared bases (green), when the mapping weights from the bases to LF1 are set to \{0,1,0.5,-1\} and those to RF1 are set to \{0,1,0.5,-1\}. This demonstrates the generation of same-phase patterns. \weightsetmeaning}}
\begin{figure}[!h]
	\centering
	\includegraphics[width=0.8\linewidth]{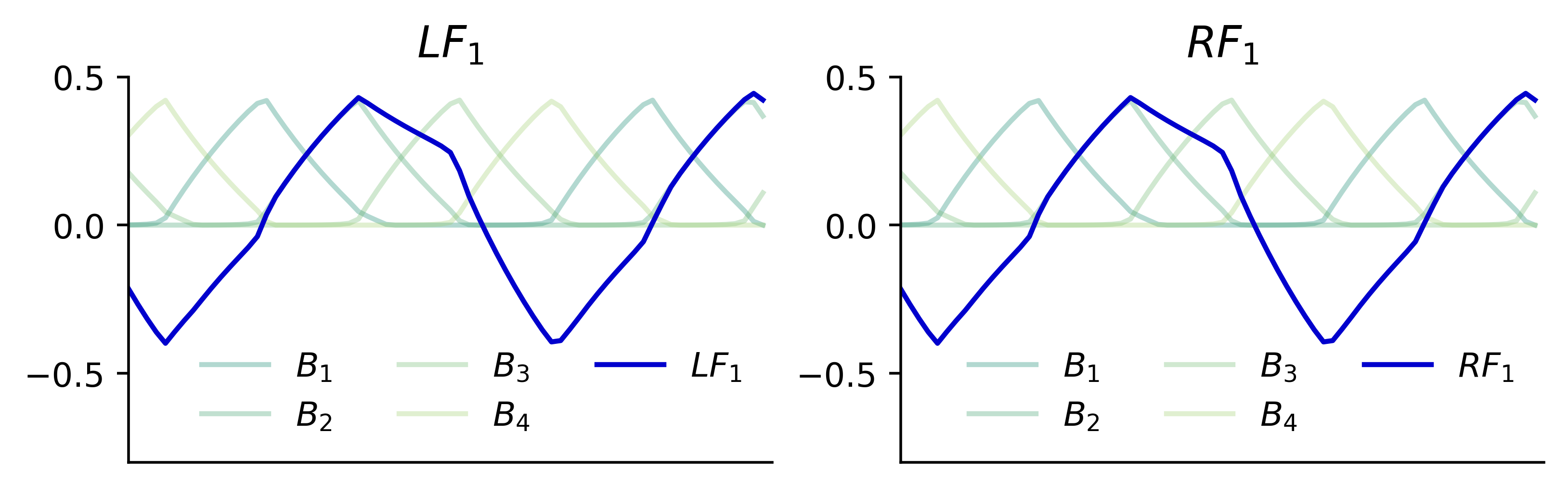}
	\caption{\legcozerocap}
	\label{fig:legco0}
\end{figure}

\def\legconinetycap{\review{(left, blue) LF1 and (right, blue) RF1 commands, presented along with their shared bases (green), when the mapping weights from the bases to LF1 are set to\{0,1,0.5,-1\} and those to RF1 are set to \{1,0.5,-1,0\}. This demonstrates the generation of 90$^\circ$-phase different patterns. \weightsetmeaning}}
\begin{figure}[!h]
	\centering
	\includegraphics[width=0.8\linewidth]{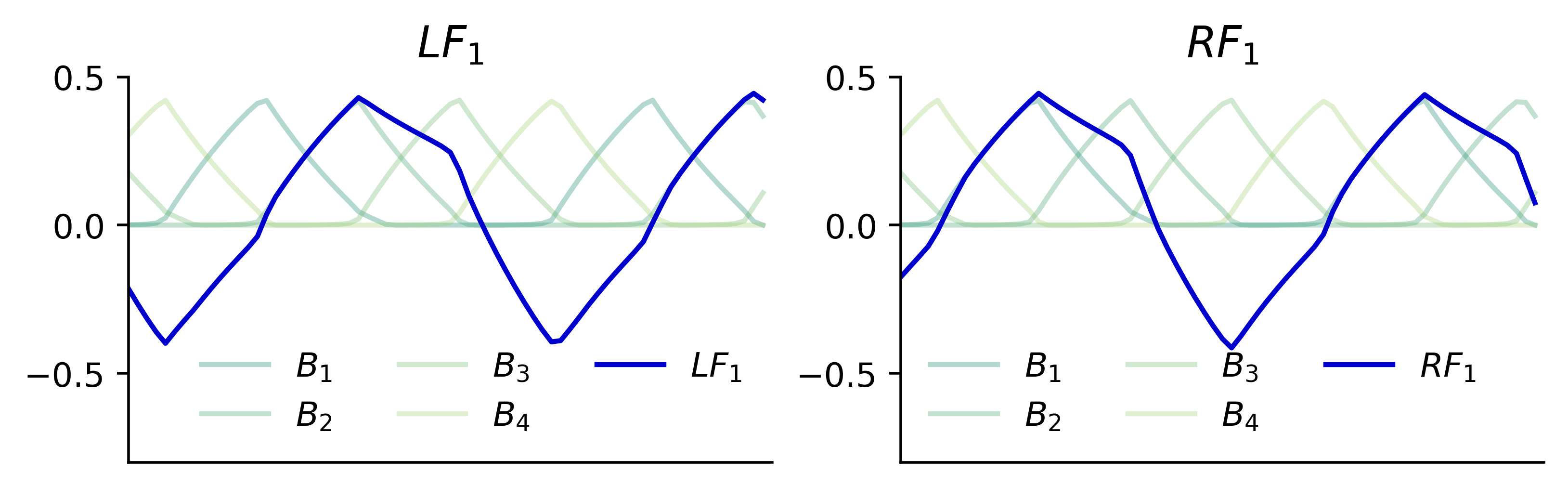}
	\caption{\legconinetycap}
	\label{fig:legco90}
\end{figure}

\def\legcoanti{\review{(left, blue) LF1 and (right, blue) RF1 commands, presented along with their shared bases (green), when the mapping weights from the bases to LF1 are set to \{0,1,0.5,-1\} but those to RF1 are set to \{-1,0,1,0.5\}. This demonstrates the generation of anti-phase patterns. \weightsetmeaning}}
\begin{figure}[!h]
	\centering
	\includegraphics[width=0.8\linewidth]{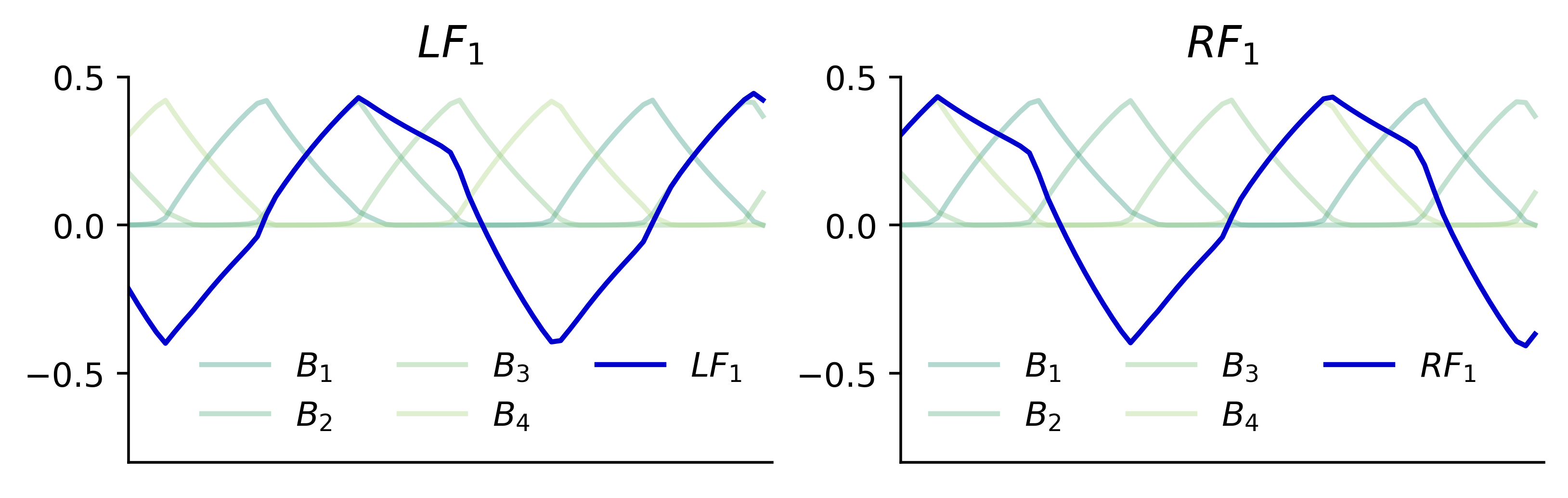}
	\caption{\legcoanti}
	\label{fig:legco180}
\end{figure}

\newpage

\section{Example of Different Leg Pattern}
\label{sec:differentshape}

\def\legdiff{\review{(blue) RF1 command, presented along with (green) the shared bases, when the mapping weights from the bases to RF1 are set to (left) \{0,1,0.5,-1\} and (right) \{0,1,-0.8,0.5\}. This demonstrates the generation of different shape patterns. Note that the mapping weights are presented as \{$w_1$, $w_2$, $w_3$, $w_4$\}, where $w_i$ denotes a mapping weight from the $i^{th}$ basis to the output: $RF_1$, and the $i^{th}$ key poses is $\approx$ 0.5$\cdot w_i$.}}
\begin{figure}[!h]
	\centering
	\includegraphics[width=0.8\linewidth]{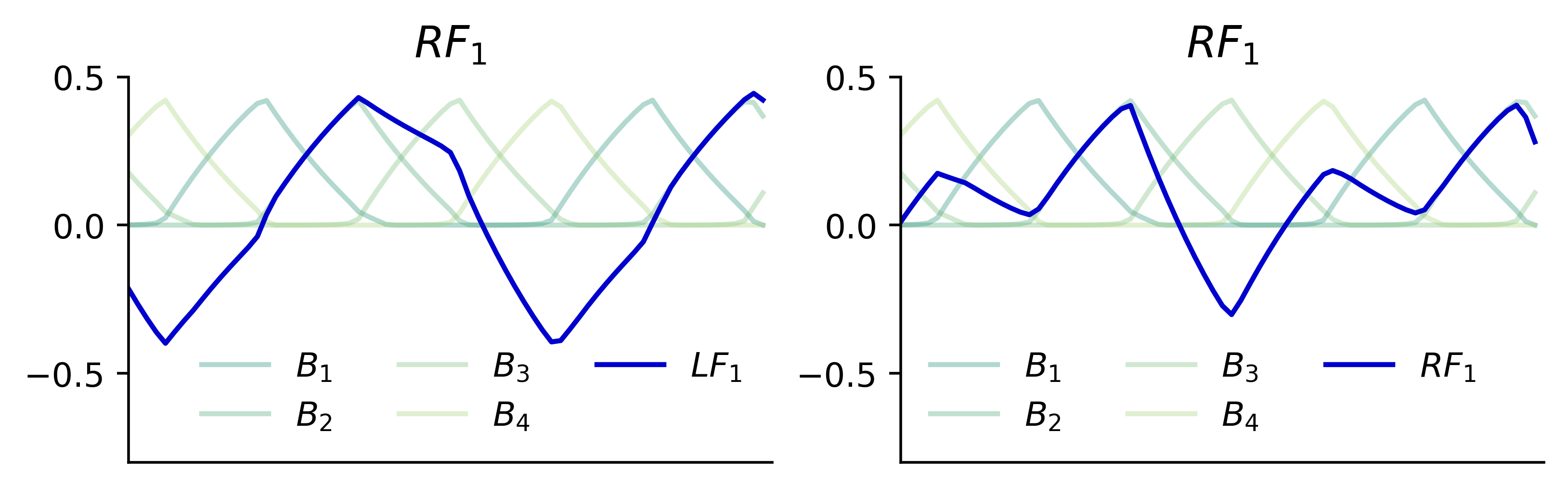}
	\caption{\legdiff}
	\label{fig:legshape}
\end{figure}

\newpage

\section{CPGRBF neural control}
\label{sec:CPGRBFfigure}

\begin{figure}[!h]
	\centering
	\includegraphics[width=0.6\linewidth]{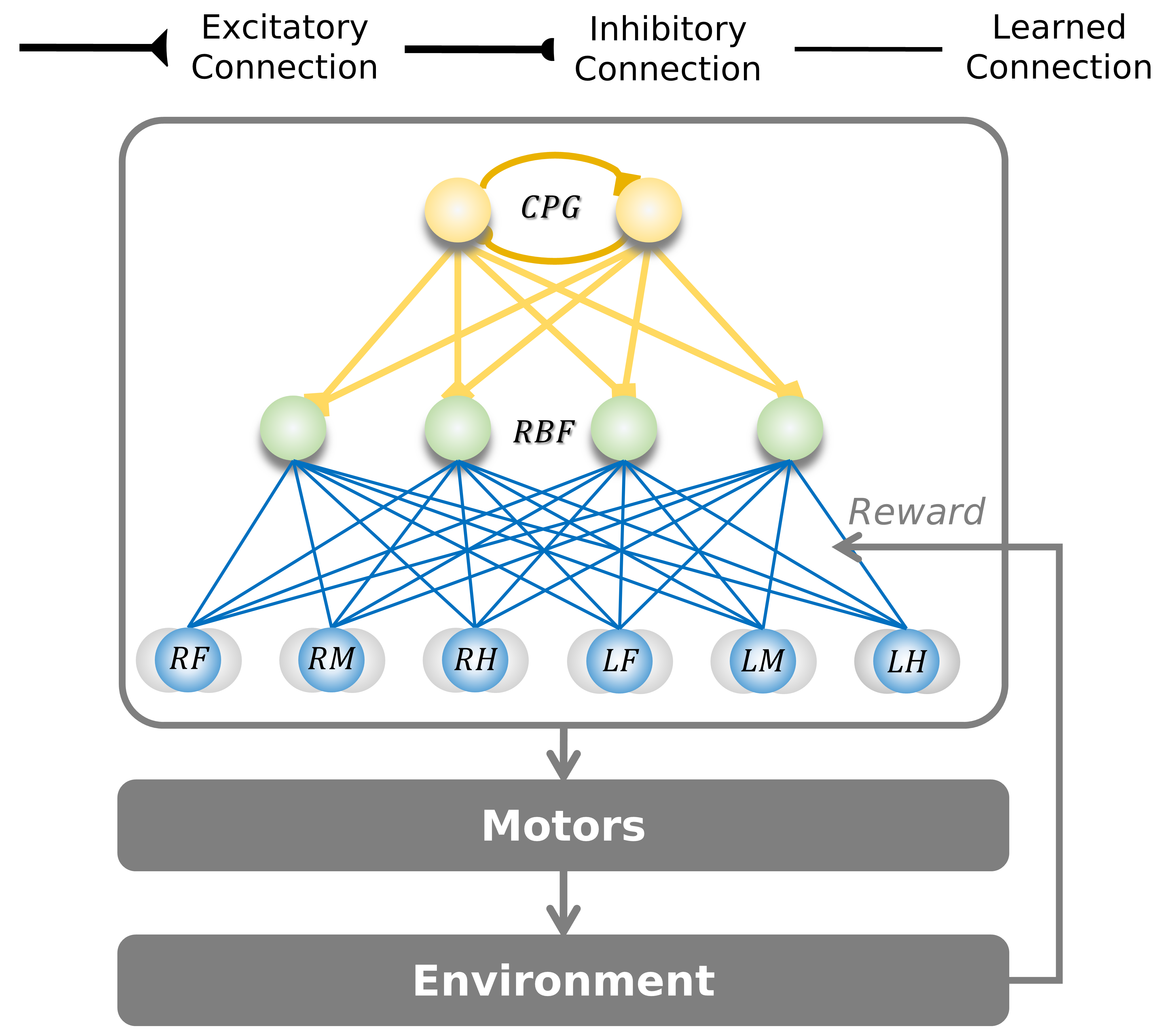}
	\caption{The structure of the CPGRBF neural control employed in the experiment, where CPG is a two-neuron oscillator network acting as the central pattern generator and RBF is a radial basis function network acting as the output mapping.}
	\label{fig:cpgrbf}
\end{figure}

\section{The Activities of Fully Connected Neural Network}
\label{sec:fcnnactivity}

\def \fcnnactcap{(a) An example of the four neural activities (i.e., $x_1$--$x_4$) of the last hidden layer of (b) a fully-connected neural network (FCNN) with 1.7k parameters and the ReLU activation function trained for 100 episodes. We can see several overlapping of neural activities within a gait cycles. This shows that \SI{17}{\percent} of each neuron activity is interfered by at least one of the others, which leads to inefficient learning in FCNN \cite{MELA,HlifeRL_optionbased,dogrobot_massivelyparallel}.}

\begin{figure}[!h]
	\centering
	\begin{subfigure}{0.6\linewidth}
		\includegraphics[width=\linewidth]{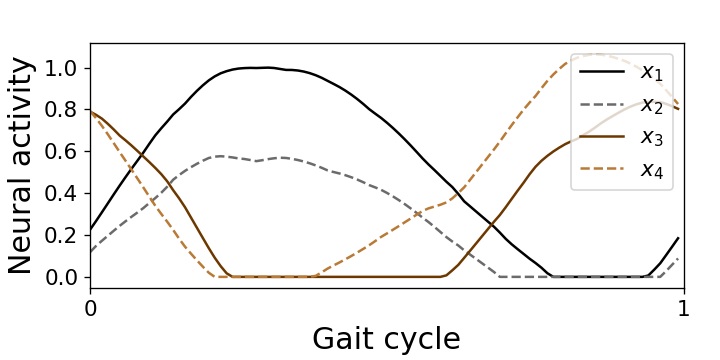}
		\caption{}
		\label{fig:fcnnactivity}
	\end{subfigure}
	\hfil
	\begin{subfigure}{0.32\linewidth}
		\includegraphics[width=\linewidth]{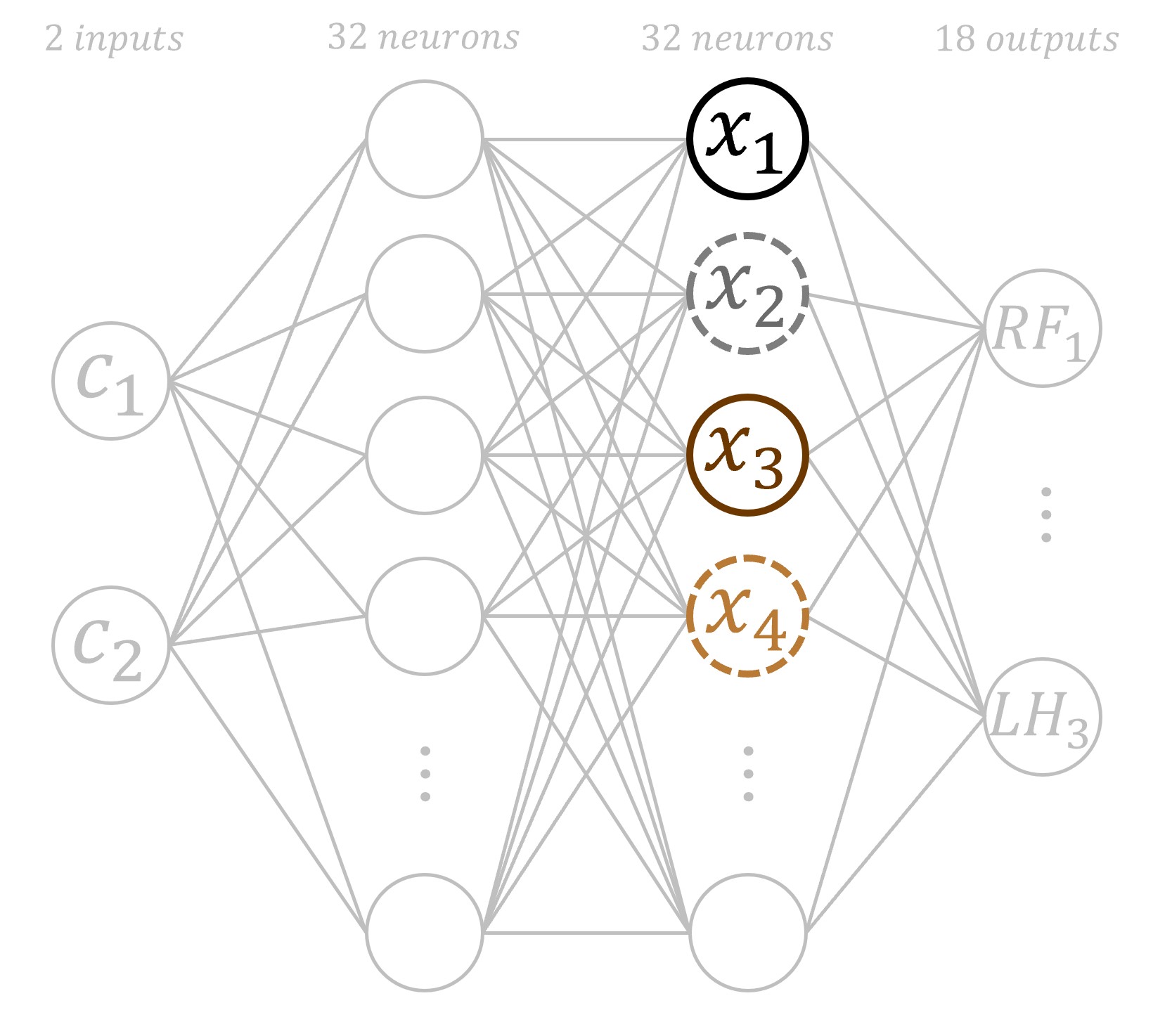}
		\caption{}
		\label{fig:fcnnnet}
	\end{subfigure}
	\caption{\fcnnactcap}%
	\label{fig:fcnn}
\end{figure}

\end{document}